\documentclass[10pt,twocolumn,letterpaper]{article}

\usepackage{cvpr}              

\usepackage{pifont}

\definecolor{checkmark}{HTML}{40826D}
\definecolor{xmark}{HTML}{E62020}

\definecolor{cvprblue}{rgb}{0.21,0.49,0.74}
\usepackage[pagebackref,breaklinks,colorlinks,allcolors=cvprblue]{hyperref}
\usepackage{booktabs,makecell,multirow,xcolor,array}
\usepackage{caption}
\usepackage{bm}

\def\paperID{22025} 
\def\confName{CVPR}
\def\confYear{2026}

\title{Are AI-Generated Driving Videos Ready for Autonomous Driving? \\ A Diagnostic Evaluation Framework}

\author{Xinhao Xiang, Abhijeet Rastogi, Jiawei Zhang\\
IFM Lab, University of California, Davis\\
{\tt\small \{xhxiang, abrastogi\}@ucdavis.edu, jiawei@ifmlab.org}
}

\begin{document}
\maketitle
\begin{abstract}

Recent text-to-video models have enabled the generation of high-resolution driving scenes from natural language prompts. These AI-generated driving videos (AIGVs) offer a low-cost, scalable alternative to real or simulator data for autonomous driving (AD). But a key question remains: can such videos reliably support training and evaluation of AD models? We present a diagnostic framework that systematically studies this question. First, we introduce a taxonomy of frequent AIGV failure modes, including visual artifacts, physically implausible motion, and violations of traffic semantics, and demonstrate their negative impact on object detection, tracking, and instance segmentation. To support this analysis, we build ADGV-Bench, a driving-focused benchmark with human quality annotations and dense labels for multiple perception tasks. We then propose ADGVE, a driving-aware evaluator that combines static semantics, temporal cues, lane obedience signals, and Vision-Language Model(VLM)-guided reasoning into a single quality score for each clip. Experiments show that blindly adding raw AIGVs can degrade perception performance, while filtering them with ADGVE consistently improves both general video quality assessment metrics and downstream AD models, and turns AIGVs into a beneficial complement to real-world data. Our study highlights both the risks and the promise of AIGVs, and provides practical tools for safely leveraging large-scale video generation in future AD pipelines.

\end{abstract}    
\section{Introduction}
\label{sec:intro}

\begin{figure}[t]
  \centering
  \includegraphics[width=\linewidth]{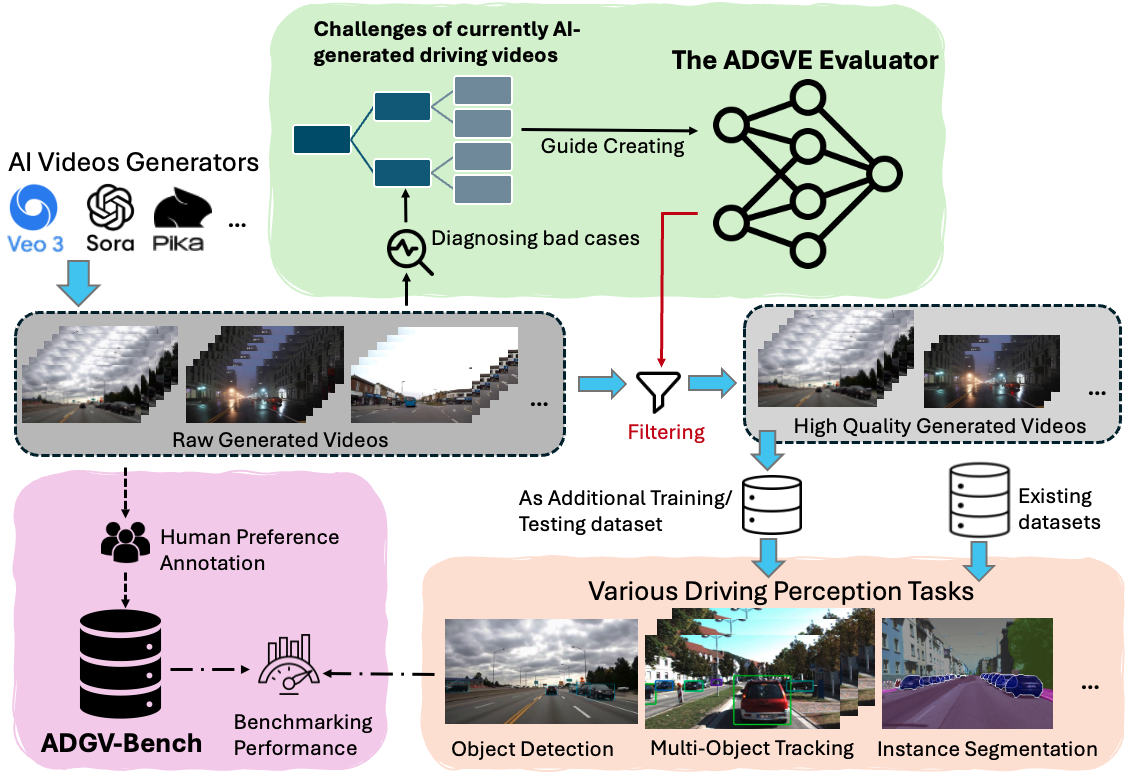}
  \caption{Overview of our pipeline. Prompt-only video generators (e.g., Veo~3, Sora, Pika) produce raw AI-generated driving videos (AIGVs). ADGVE diagnoses failure cases and filters out low-quality clips, yielding a high-quality subset that could combine with existing real datasets as additional training/testing data for downstream driving perception tasks such as object detection, multi-object tracking, and instance segmentation).}
  \label{fig:pipeline}
\end{figure}

\begin{table*}[t]
\centering
\resizebox{\linewidth}{!}{
\begin{tabular}{lcccc}
\toprule
Criterion & Real Data~\cite{KITTI, nuScenes, WOD, Argoverse2} & Simulator~\cite{CARLA} & AD-Specific Gen~\cite{Panacea+, DriveGenVLM, VaViM_VaVAM} & \textbf{Prompt-Only AIGV}~\cite{sora2024, pika, Veo3, HunyuanVideo} \\
\midrule
Cost \& setup time        & Very high & Medium & High & Low \\
No-code usability         & Low       & Low    & Low--Medium & High \\
Edge-case flexibility / control & Low & Medium (scripted) & High & Very high \\
Zero-shot / open-world composition & Low & Low & Medium & Very high \\
\bottomrule
\end{tabular}
}
\caption{Comparison of video generation paradigms for AD data synthesis. Prompt-only AIGVs offer uniquely low cost, high usability, and strong coverage for edge cases and open-world compositions, but require dedicated quality control (this work's focus) to be safely integrated into AD pipelines.}
\label{tab:generation_paradigms}
\end{table*}

Autonomous driving (AD) demands large-scale, diverse, and high-quality video data to train and evaluate perception models. Traditional pipelines rely on real-world fleet data or game-engine-based simulators such as CARLA~\cite{CARLA}, which require substantial engineering effort and are still limited in diversity or controllability. Recent text-to-video models~\cite{sora2024,pika} make it possible to synthesize high-resolution driving scenes directly from prompts. These prompt-only AI-generated driving videos (AIGVs) are cheap to produce, easy to script, and can in principle cover rare edge cases and open-world compositions at scale. Table~\ref{tab:generation_paradigms} contrasts prompt-only AIGVs with real data, simulators, and AD-specific video generators.

However, a critical question remains: \textbf{Can these AI-generated videos be safely used for training or evaluating real-world AD models?} Existing video generators are not optimized for the structural, semantic, and behavioral correctness required in traffic scenes. In practice, AIGVs frequently exhibit visual artifacts, physically impossible motion, lane and rule violations, or inconsistent object behaviors. Naively mixing such clips into training or evaluation can inject harmful noise, but current benchmarks and generic video quality assessment (VQA) tools are not designed to detect these AD-specific deficiencies.

To address this gap, we analyze the \emph{challenges}, \emph{impacts}, and \emph{mitigation strategies} of using AIGVs for driving perception. We first construct a taxonomy of failure modes along two axes—\emph{Spatiotemporal Realism} and \emph{Semantic \& Behavioral Realism}—and show how they manifest in videos from modern prompt-only generators. We then build \textbf{ADGV-Bench}, a driving-focused benchmark that couples human quality annotations with downstream tasks such as object detection, multi-object tracking, and instance segmentation. On top of this, we develop \textbf{ADGVE}, a driving-aware evaluator that combines static cues, temporal consistency, lane-obedience checks, and VLM-guided reasoning into a single quality score. As illustrated in Fig.~\ref{fig:pipeline}, ADGVE filters raw AIGVs into a high-quality subset that can be used to benchmark video quality and to augment real datasets for perception training.
Our contributions are summarized as follows:
\begin{enumerate}
    \item We provide, to our knowledge, the first structured taxonomy of quality challenges specific to prompt-only AIGVs in autonomous driving.
    \item We introduce ADGV-Bench, a benchmark with human preference labels and multi-task perception annotations tailored to evaluating AIGVs for AD.
    \item We propose ADGVE, a driving-aware quality evaluator, and show through extensive experiments that its filtering both aligns with existing VQA/AIGVE metrics and consistently improves detection, tracking, and segmentation when AIGVs are used to augment real-world datasets.
\end{enumerate}

\section{Related Work}
\label{sec:related}

\subsection{AI-Generated Driving Videos}
Recent work has begun to explore AI-generated videos as a source of driving data.  
Panacea+~\cite{Panacea+}, DriveGenVLM~\cite{DriveGenVLM}, and VaViM / VaVAM~\cite{VaViM_VaVAM} generate AD scenes by conditioning on structured inputs such as HD maps, bird’s-eye semantics, or trajectory priors, achieving strong scene control but requiring domain-specific assets and engineering.  
In contrast, general text-to-video models such as Sora~\cite{sora2024}, HunyuanVideo~\cite{HunyuanVideo}, and Pika~\cite{pika} enable prompt-only generation directly from natural language.  
As summarized in Table~\ref{tab:generation_paradigms}, prompt-only AIGVs offer uniquely low cost, no-code usability, and broad open-world coverage, making them attractive for scalable AD data synthesis—provided their quality can be reliably assessed.

\subsection{AI Generated Videos Evaluation (AIGVE)}

Evaluating AI-generated video quality is an active research area~\cite{liu2024surveyaigve}.  
Classical metrics such as FID and FVD~\cite{FID,FVD} measure distributional similarity but are insensitive to many temporal or physics-driven failures.  
Recent AIGV benchmarks, including AIGVE-Tool~\cite{AIGVE_Tool}, introduce richer probes for temporal stability and visual coherence, while VQA models such as LightVQA+~\cite{Light_VQA_plus} and GSTVQA~\cite{GSTVQA} provide general-purpose perceptual scores.  
However, these efforts are largely domain-agnostic and do not explicitly enforce traffic rules, lane obedience, or physically plausible ego motion.  
Our work complements this line by focusing on prompt-only \emph{driving} videos and by designing an evaluator, ADGVE, whose score is explicitly aligned with AD-specific constraints and downstream perception performance.


\subsection{Dataset Quality and Safety in AD Training}

Real-world AD datasets such as nuScenes~\cite{nuScenes}, Argoverse~2~\cite{Argoverse2}, and KITTI~\cite{KITTI} emphasize sensor fidelity and annotation accuracy, while simulators like CARLA~\cite{CARLA} and MetaDrive~\cite{MetaDrive} offer controllable synthetic data with perfect labels.  
These sources, however, are expensive to scale and do not directly address the new failure modes introduced by generative video models.  
Low-quality AIGVs can inject unrealistic motion patterns, distorted objects, and semantically incorrect behaviors into training, potentially degrading model safety.  
Prior work rarely quantifies how such generative artifacts propagate to downstream AD performance.  
Our study fills this gap by systematically diagnosing the impact of prompt-only AIGVs on detection, tracking, and segmentation, and by proposing ADGV-Bench and ADGVE as a quality-aware pipeline for filtering harmful samples before they are used in AV training or evaluation.

\section{Method}
\label{sec:method}

\subsection{Diagnosing the Challenges of AI-Generated Driving Videos}
\label{sec:challenges}

\begin{figure}[t]
  \centering
  \includegraphics[width=\linewidth]{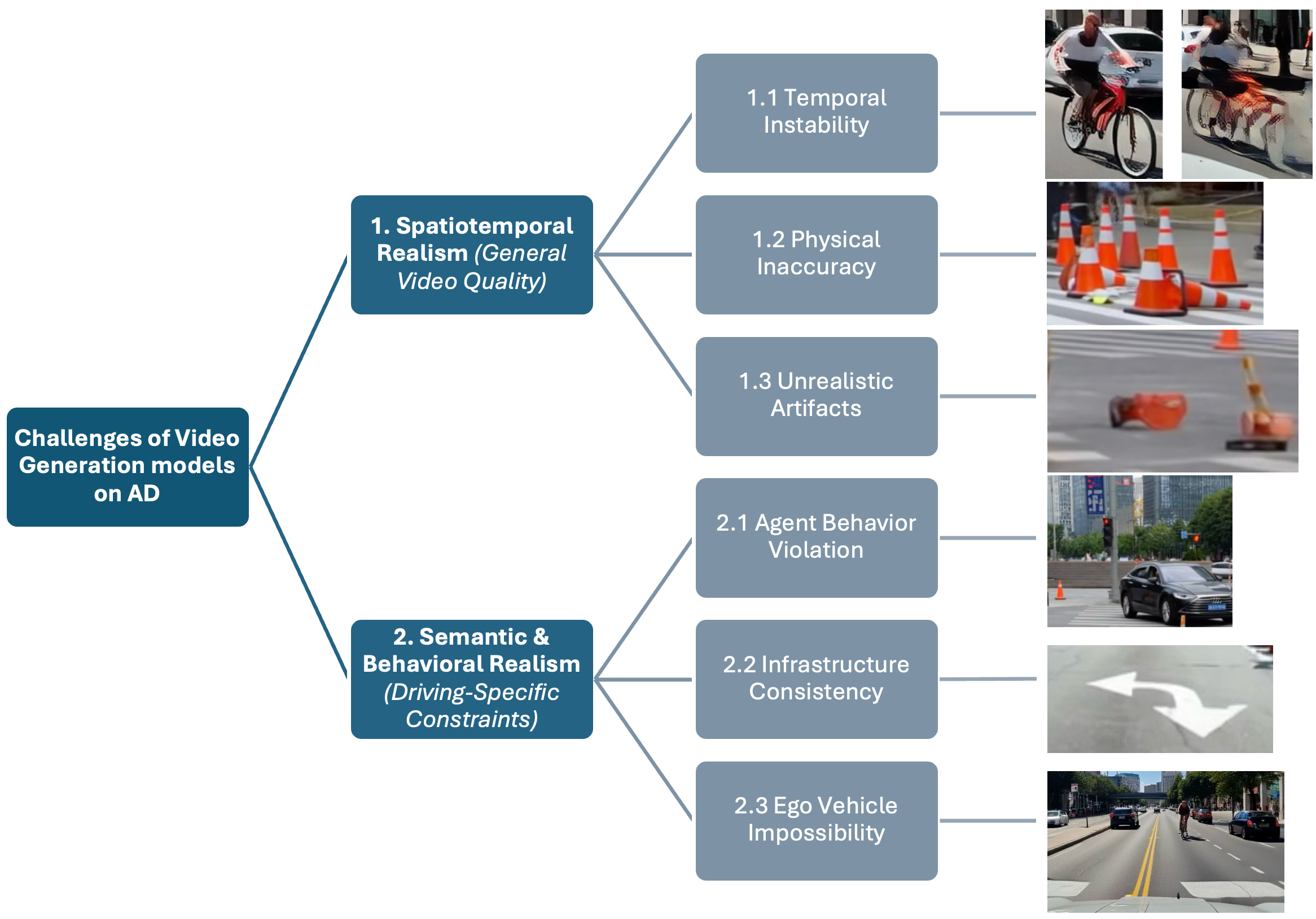}
  \caption{Challenges of currently AI-generated driving videos, with example failure cases. More failure cases are provided in Supp.~\ref{sec:challenges_more}.}
  \label{fig:challenges}
\end{figure}

\begin{figure*}[t]
  \centering
  \includegraphics[width=\linewidth]{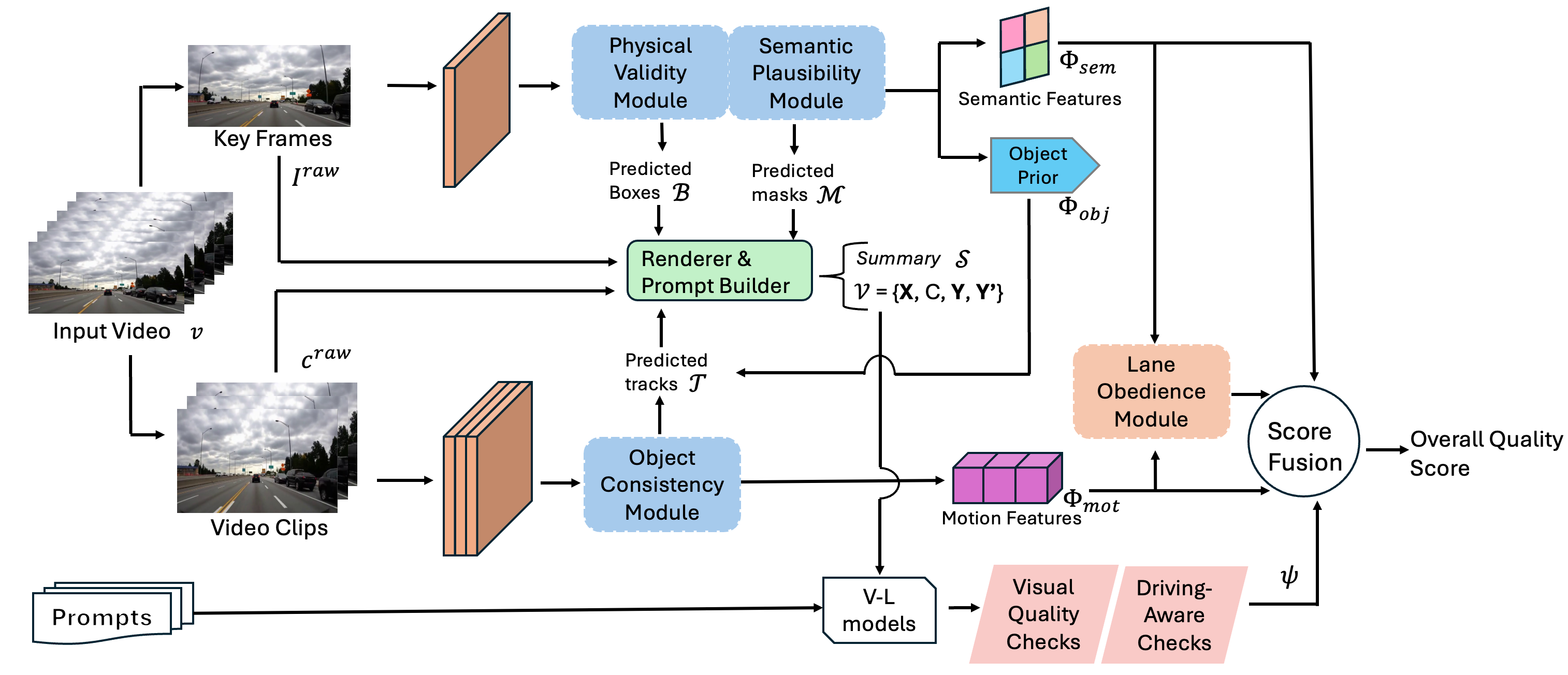}
  \caption{Overview of the ADGVE evaluator. Given an AI-generated driving video $v$, ADGVE extracts static object and infrastructure priors (Physical Validity, Semantic Plausibility), temporal tracks and motion features (Object Consistency), and converts them into a visual bundle $\mathcal{V} = \{\mathbf{X}, \mathbf{C}, \mathbf{Y}, \mathbf{Y}'\}$ plus textual summaries $\mathcal{S}$ for a video-language model. A lane-obedience module computes an additional geometric score from lanes and trajectories. All VLM-derived scores and descriptors are fused into a single driving-aware quality score $S_{\text{overall}}$, which we use to filter low-quality AIGVs (default threshold $S_{\text{overall}} > 0.2$) and to analyze their impact on downstream driving perception.}
  \label{fig:adgve}
\end{figure*}



Before using AI-generated driving videos (AIGVs) as training or evaluation data for autonomous driving (AD), we first diagnose their typical failure modes. Despite recent progress in generative video models, AIGVs still exhibit subtle but systematic artifacts that can be harmful for downstream perception~\cite{liu2024surveyaigve}. We inspect a large number of traffic videos generated by leading models such as Pika~\cite{pika} and Sora~\cite{sora2024} and conduct qualitative error analysis. Based on these observations, we propose a taxonomy of challenges for AI-generated traffic scenes (Fig.~\ref{fig:challenges}), organized along two axes: \textbf{Spatiotemporal Realism} and \textbf{Semantic \& Behavioral Realism}. These axes capture both generic video fidelity and driving-specific correctness.

\paragraph{Spatiotemporal Realism.}

This axis measures whether a video is temporally coherent and physically plausible. Failures here can harm perception even before considering traffic semantics.
\emph{Temporal Instability (1.1)} denotes objects whose appearance or geometry changes abruptly across frames (e.g., a cyclist whose bike intermittently disappears), which breaks object permanence and degrades tracking or temporal fusion. 
\emph{Physical Inaccuracy (1.2)} captures violations of basic physics or geometry, such as “melting’’ cars, misaligned cones, or warped horizons; such distortions conflict with inductive biases in detection and motion forecasting models. 
\emph{Unrealistic Artifacts (1.3)} refers to hallucinated or nonsensical objects that do not exist in real traffic scenes, which can corrupt the semantic structure of the environment and mislead recognition or grounding.


\paragraph{Semantic \& Behavioral Realism.}

This axis focuses on whether the video respects driving rules, behavior priors, and infrastructure logic. Low realism here can distort the learning targets of AD systems.
\emph{Agent Behavior Violation (2.1)} covers non-ego agents (\textit{e.g.} cars, pedestrians, cyclists) that move in implausible ways, such as cars sliding sideways, drifting across double solid lines, or pedestrians walking in the middle of the road, which mis-shape the behavior distributions that prediction and risk assessment rely on. 
\emph{Infrastructure Consistency (2.2)} targets errors in static road elements, including contradictory lane markings, missing or unlit traffic lights, or signs that conflict with the scene; these errors undermine map-based localization and rule-based reasoning. 
\emph{Ego Vehicle Impossibility (2.3)} describes clips where the ego car itself violates basic constraints, \textit{e.g.}, occupying multiple lanes, driving off-road, or exhibiting jerky, non-smooth motion, which breaks assumptions used by lane segmentation and trajectory tracking.

\paragraph{From taxonomy to evaluator design.}

This taxonomy directly guides the design of our ADGVE evaluator (Sec.~\ref{sec:adgve}): temporal instability and physical inaccuracy motivate modules for object consistency and physical validity, while agent, infrastructure, and ego violations inform lane-obedience and semantic-plausibility checks. By grounding ADGVE in these concrete categories, we ensure that its scores target exactly the failure patterns that most strongly affect downstream AD perception, setting the stage for our multi-task evaluation in Sec.~\ref{sec:exp}.



\subsection{The ADGVE Evaluator}
\label{sec:adgve}

Guided by the taxonomy in Sec.~\ref{sec:challenges}, we design \textbf{ADGVE}, a driving-aware evaluator that assigns each AI-generated driving video a scalar quality score together with interpretable sub-scores. Given a video
$v = \{I_t\}_{t=1}^{T}$, ADGVE operates at two complementary granularities: (i) \textbf{key frames} that capture static appearance and scene layout, and (ii) short \textbf{video clips} that capture temporal dynamics. As illustrated in Fig.~\ref{fig:adgve}, static branches extract object and infrastructure priors, a temporal branch estimates tracks and motion, a Renderer \& Prompt Builder converts these priors into VLM-ready inputs and summaries, and a lane-obedience module adds geometric rule checks. All cues are finally fused into an overall driving-aware score $S_{\text{overall}} \in [0,1]$, which we later use for filtering. Unless otherwise stated, we mark clips with $S_{\text{overall}} > 0.2$ as ``high-quality'' and treat them as candidates.

We split each video into $M=8$ equal-length clips $\{c_m\}_{m=1}^{M}$ and take the center frame of each as a key frame $\{I_{f_k}\}_{k=1}^{K}$, with $K=M$. Key frames are routed to the \textbf{Physical Validity} and \textbf{Semantic Plausibility} modules, while clips go to the \textbf{Object Consistency} module. 
The resulting priors---bounding boxes $\mathcal{B}$, semantic masks $\mathcal{M}$, and tracklets $\mathcal{T}$---are rendered into a visual bundle
$\mathcal{V} = \{\mathbf{X}, \mathbf{C}, \mathbf{Y}, \mathbf{Y}'\}$, where $\mathbf{X}$, $\mathbf{C}$, $\mathbf{Y}$, and $\mathbf{Y}'$ denote key-frame triplets, ROI crops, clip pairs, and multi-level sub-clips respectively (detailed below), and are summarized into text $\mathcal{S}$ for a video-language model (VLM).
In parallel, a rule-based \textbf{Lane Obedience} module computes a geometric score $s_{\text{lane}}$. We next detail each component.

\paragraph{Static Perception.}

We operate on the $K$ key frames and extract two complementary static cues: explicit priors (boxes, masks) that can be rendered, and compact tokens that summarize the scene for fusion.
The \textbf{Physical Validity} module is a lightweight detector for traffic participants (vehicle/pedestrian/cyclist). For each key frame $I_{f_k}$ it returns
$\mathcal{B}_k = \{(b_i, c_i, p_i)\}_{i=1}^{N_k}$ after class-wise NMS, with $b_i \in \mathbb{R}^4$ bounding boxes, $c_i$ class labels, and $p_i \in [0,1]$ confidences. For each detection we pool a $d$-dimensional feature from the detector’s final feature map (RoIAlign followed by global average pooling), forming the object-prior tokens $\Phi_{\text{obj}}$; the tuples $\mathcal{B}$ are kept for rendering and prompt construction.
The \textbf{Semantic Plausibility} module is a semantic segmenter for road infrastructure (road/sidewalk/lane/crosswalk/sign/light). It outputs per-frame masks
$\mathcal{M}_k = \{(m_\ell, t_\ell, q_\ell)\}_{\ell=1}^{L_k}$ with pixel sets $m_\ell$, classes $t_\ell$, and confidences $q_\ell \in [0,1]$. For each semantic class we perform masked average pooling over the segmenter’s feature map to obtain semantic tokens $\Phi_{\text{sem}}$. The masks $\mathcal{M}$ support both rendering and lane-obedience analysis, while $(\Phi_{\text{obj}}, \Phi_{\text{sem}})$ jointly encode \emph{what is where} in the scene.

\paragraph{Temporal Perception.}

The \textbf{Object Consistency} module processes the $M$ clips using a lightweight video backbone and tracker. For each clip $c_m$ it links detections into short tracklets $\mathcal{T} = \{\tau_j\}$, where
$\tau_j = \{(b_j^{t}, p_j^{t})\}_t$ are time-ordered boxes with confidences. From these tracks we derive basic kinematics, such as speed, heading change, and lateral drift. The backbone features are temporally pooled into a motion descriptor $\Phi_{\text{mot}}$; when tracking confidence is low we optionally add optical flow features $F$. Tracklets $\mathcal{T}$ are later rendered as polylines for VLM queries (e.g., smoothness, lane centering, solid-line crossings), while $\Phi_{\text{mot}}$ serves as a dense temporal descriptor for fusion.


\paragraph{Renderer \& Prompt Builder.}

Given boxes $\mathcal{B}$, masks $\mathcal{M}$, and tracklets $\mathcal{T}$, we convert them into VLM-ready visual inputs and compact text summaries.

\textit{Rendering.}
Each VLM query receives paired visual evidence: a raw view and an overlaid view built from priors. For every key frame $I_{f_k}$ we form a triplet
$\mathbf{X}_k = \{I_{f_k}^{\text{raw}}, I_{f_k}^{\text{B}}, I_{f_k}^{\text{M}}\}$ consisting of the original image, the image with boxes $\mathcal{B}_k$, and the image with masks $\mathcal{M}_k$. For each clip $c_m$ we form a pair $\mathbf{Y}_m = \{c_m^{\text{raw}}, c_m^{\text{T}}\}$, where $c_m^{\text{T}}$ overlays temporally faded track polylines from $\mathcal{T}$ (and flow arrows when available). We additionally create ROI crops $\mathbf{C}$ for fine-grained checks (vehicles, pedestrians, cyclists, traffic lights/signs, lane arrows/markings) by expanding each box with a margin, clipping to image bounds, and letterboxing to the VLM input size. For temporal hierarchy, we also follow LightVQA+~\cite{Light_VQA_plus} to extract multi-level sub-clips $\mathbf{Y}'_m$ from each $c_m$. All frames, clips, sub-clips, and crops collectively form the visual bundle
$\mathcal{V} = \{\mathbf{X}, \mathbf{C}, \mathbf{Y}, \mathbf{Y}'\}$.

\textit{Prompt building.}
For each element of $\mathcal{V}$ we attach a concise textual summary $\mathcal{S}$ distilled from $\mathcal{B}$, $\mathcal{M}$, and $\mathcal{T}$. The summary reports object counts per class, coarse spatial layout on a left–center–right by near–mid–far grid, approximate size categories, lane attributes (solid/dashed/double-solid, crosswalks, arrows, signals), and short motion descriptors (direction, lateral drift, speed regime, smoothness). Continuous quantities are discretized into semantic bins to absorb estimation noise. The VLM receives the rendered image or clip as visual input and a prompt template with the injected summary $\mathcal{S}$ as text input. For each check, we formulate multiple-choice questions and convert the resulting answer probabilities into calibrated features for downstream fusion (the full prompt set is listed in Supp.~\ref{subsection:full_set}).

\paragraph{Vision–Language Guided Checks.}

Given $(\mathcal{V}, \mathcal{S})$, we query a video-language model to obtain two families of scores: (i) \emph{visual-quality checks}, covering generic degradations such as brightness, noise, blur, and exposure flicker, and (ii) \emph{driving-aware checks}, covering traffic semantics and rules such as lane-centering, solid-line and crosswalk compliance, signal obedience, wrong-way or sidewalk driving, motion smoothness, and right-of-way. Following LightVQA+~\cite{Light_VQA_plus}, we use a \emph{key-frame path} that operates on $\mathbf{X}$ and ROI crops for spatial judgments, and a \emph{multi-level clip path} that operates on $\mathbf{Y}$ and $\mathbf{Y}'$ for temporal judgments. The VLM outputs a compact vector
$\boldsymbol{\psi} = [\boldsymbol{\psi}^{\text{frame}}; \boldsymbol{\psi}^{\text{clip}}]$ of calibrated scores, where
$\boldsymbol{\psi}^{\text{frame}}$ aggregates key-frame checks and
$\boldsymbol{\psi}^{\text{clip}}$ aggregates temporal consistency and behavior checks.

\paragraph{Lane Obedience Module.}

From semantic masks $\mathcal{M}$ we extract lane centerlines and solid vs.\ dashed boundaries; from tracklets $\mathcal{T}$ we obtain per-agent trajectories. A rule-based scorer $h(\mathcal{M}, \mathcal{T})$ computes a lane-obedience score $s_{\text{lane}} \in [0,1]$ by combining three normalized components: (i) lane-centering error (mean distance of vehicle tracks to the nearest lane centerline), (ii) solid-line violation rate (fraction of polyline segments crossing solid boundaries), and (iii) crosswalk yielding/encroachment penalties. Weights for these components are fixed on a dev set; implementation details are in Supp.~\ref{subsection:lane_obedience}.


\paragraph{Score Fusion.}

After obtaining key–frame scores $\boldsymbol{\psi}^{\text{frame}}$ and clip-level scores
$\{s^{\text{clip}}_{m}\}_{m=1}^{M}$ from the multi-level path, we first aggregate the temporal scores with an HVS-style attention~\cite{HVS} and then fuse all cues.
Each clip $c_m$ is summarized as
$r_m = [\,\mathrm{GAP}(\Phi_{\text{mot}}^{\,m}),\; s^{\text{clip}}_{m}\,]$,
where $\mathrm{GAP}(\cdot)$ denotes global average pooling and $\Phi_{\text{mot}}^{\,m}$ is the motion feature of $c_m$. We compute attention weights:
\begin{equation}
\alpha_m = \frac{\exp(u^\top r_m)}{\sum_{r=1}^{M} \exp(u^\top r_r)},
\qquad
S_{\text{clip}} = \sum_{m=1}^{M} \alpha_m\, s^{\text{clip}}_{m},
\end{equation}
where $u \in \mathbb{R}^{\dim(r_m)}$ is a learnable attention vector, optimized together with the fusion regressor.

Finally, we predict an overall score by combining the VLM features, the HVS-aggregated temporal score, the lane-obedience score, and pooled static/temporal descriptors via a lightweight regressor:
\begin{equation}
\begin{aligned}
S_{\text{overall}}
&= \sigma\!\bigg(
    w^\top\!\big[
      \psi^{\text{frame}},\,
      S_{\text{clip}},\,
      s_{\text{lane}},\\
&\qquad
      \mathrm{GAP}(\Phi_{\text{obj}}),\,
      \mathrm{GAP}(\Phi_{\text{sem}}),\,
      \mathrm{GAP}(\Phi_{\text{mot}})
    \big]\bigg)
\end{aligned}
\end{equation}
where $w$ is a learnable weight vector and $\sigma(\cdot)$ maps to $[0,1]$.
During training, the fusion parameters $\{u,w\}$ are optimized with a combination of mean absolute error between $S_{\text{overall}}$ and human quality scores and a RankNet-style pairwise ranking loss on clip pairs, following~\cite{Light_VQA_plus}; at test time we only use the predicted scalar $S_{\text{overall}}$.

\subsection{ADGV-Bench Benchmark Dataset}
\label{subsec:ADGV_bench}

\begin{figure}[t]
  \centering
  \includegraphics[width=\linewidth]{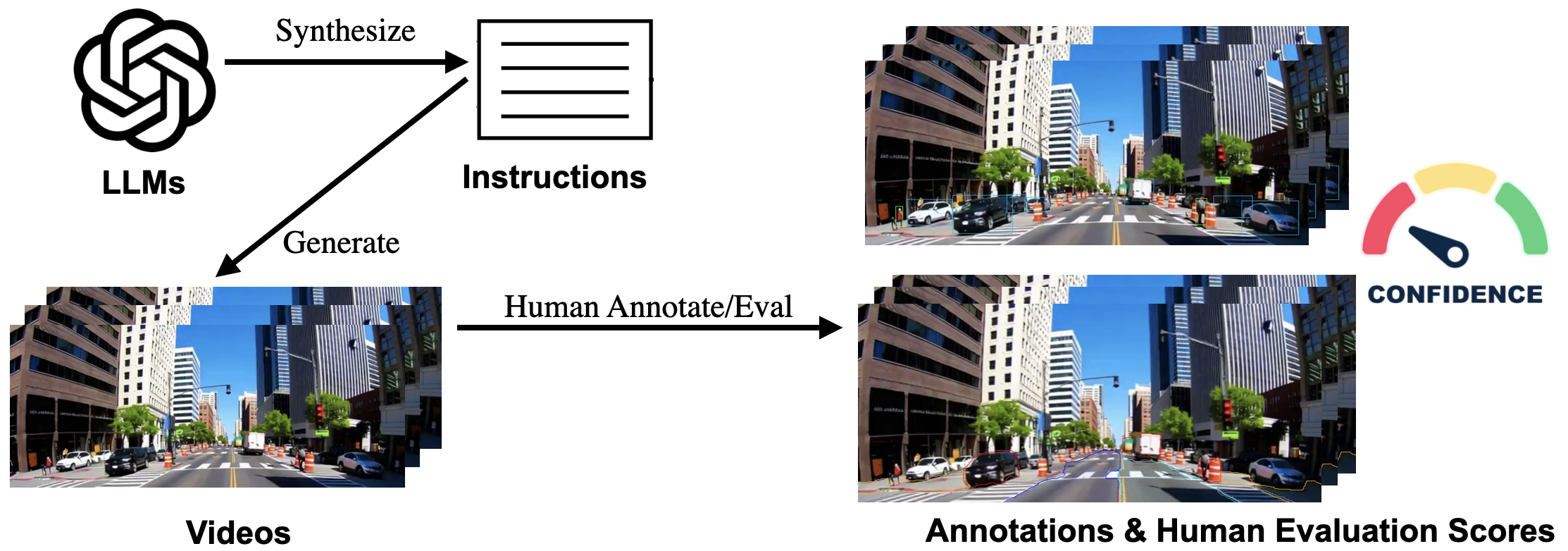}
  \caption{ADGV-Bench collection process. We first use LLM to create prompts, then use these prompts to get AI-generated driving videos from various video generator models. After that we do manual-evaluation about each quality of video, getting quality scores, and annotate each frames with bounding boxes, tracks and masks under selected traffic related categories.}
  \label{fig:data_gen}
\end{figure}


To systematically study the failure modes identified in Sec.~\ref{sec:challenges} and to fairly evaluate the ADGVE evaluator in Sec.~\ref{sec:adgve}, we require a driving–focused benchmark that (i) exercises diverse scene factors and temporal dynamics, (ii) is generator–agnostic yet comparable across models, and (iii) carries perception labels to quantify downstream utility. 
Existing video VQA or generic AIGV datasets lack traffic–rule semantics, such as lane structure and multi–object temporal annotations, making them not suitable for our goals. 
We therefore build \textbf{ADGV-Bench}: a prompt–driven, model–agnostic benchmark that pairs instructions, AI-generated driving video clips, and dense perception annotations so we can test both visual quality and driving–aware constraints and analyze how quality impacts core AD tasks.
We summarize the current standard data collection steps for ADGV-Bench in Fig.~\ref{fig:data_gen}. 


\begin{table}[t]
\centering
\small
\setlength{\tabcolsep}{6pt}
\begin{tabular}{l l}
\toprule
\textbf{Factor Dimension} & \textbf{Examples (subset)} \\
\midrule
Environment ($E$) & urban street, highway, roundabout \\
Weather ($W$) & overcast, rain, night \\
Ego behavior ($B$) & lane change, left turn, stop-and-go \\
Scene dynamics ($D$) & roadwork cones, cyclist in lane, jaywalker \\
\bottomrule
\end{tabular}
\caption{Subset of vocabulary used to compose prompts. 
The full candidate lists and sampling priors are in Supp.~\ref{sec:instructions}.}
\label{tab:prompt_dim}
\end{table}

\paragraph{Prompt Instruction Generation.}
We build video-generation instructions (i.e., text prompts) in two complementary regimes so the benchmark spans both controlled coverage and open-ended diversity.

\textit{Structured prompts.}
We generate structured prompts from a small bank of slot–filling templates
$\mathcal{T}=\{\tau_1,\tau_2,\dots\}$ rather than a single fixed pattern.
Each template composes the four factors—Environment ($E$), Weather ($W$),
Ego behavior ($B$), and Scene dynamics ($D$)—in varied orders and with diverse
connectives/surface forms to avoid repetitive phrasing and to match a target length
distribution. One possible template is:
\emph{“In the driver’s front–camera view of a $\{E\}$ under $\{W\}$, the ego vehicle
$\{B\}$ while the scene shows $\{D\}$.”}
Other templates paraphrase and reorder the slots, e.g.,
\emph{“Under $\{W\}$ in a $\{E\}$ scene, the ego car $\{B\}$ as $\{D\}$ unfolds,”}
or
\emph{“$\{D\}$ occurs in a $\{E\}$ setting; with $\{W\}$ conditions, the ego vehicle $\{B\}$.”}
For each slot we sample from curated vocabularies (Table~\ref{tab:prompt_dim} show a subset of the vocabularies we used),
apply synonym/phrase variants, and enforce de–duplication by lexical and embedding
similarity thresholds.


\textit{Free-style prompts.}
To stress open-world composition, we let the LLM describe driving scenes while conditioning on $(E,W,B,D)$ as soft hints.
We keep the same length bins and filtering rules, and reject outputs that collapse back to the structured template.
This produces diverse phrasings and rare combinations that the structured set may miss.x
To further control quality, we run a light post-edit pass to fix formatting and resolve contradictions (e.g., ``clear night sun’’). The complete template set and sampling priors, a detailed description of each of the four factors, as well as more prompts examples are provided in Supp.~\ref{sec:instructions}.

\begin{table}[t]
\centering
\setlength{\tabcolsep}{8pt}
\resizebox{0.8\linewidth}{!}{
\begin{tabular}{lccc}
\toprule
\textbf{Model} & \textbf{Resolution} & \textbf{Frame rate} & \textbf{Duration} \\
\midrule
Veo\,3~\cite{Veo3}  & 1920$\times$1080 & 24 fps & 5s \\
Pika~\cite{pika}    & 1280$\times$720      & 24 fps & 5s \\
Sora~\cite{sora2024}    & 854$\times$480         & 24 fps & 5s \\
\bottomrule
\end{tabular}
}
\caption{Specifications of generated videos for each model.  }
\label{tab:video_gen}
\end{table}

\paragraph{Video Generation.}
Each instruction prompt will be used to render AI-generated driving videos by three SOTA prompt-only video generators: Veo 3~\cite{Veo3}, Pika~\cite{pika}, and Sora~\cite{sora2024}.  
To make cross\textendash model comparison fair while still exposing long\textendash range temporal effects, we fix the clip length to 5s for all models.  
For every video generator, we request its highest quality settings permitted by our compute/budget and the model’s public API limits, such as resolution, frame rate, and quality preset.  
Table~\ref{tab:video_gen} summarizes the specifications of our generated videos from each video generator.

\paragraph{Video Annotation.}
For each video, we annotate 2D object bounding boxes and tracks for three traffic actors: \textit{vehicle}, \textit{pedestrian}, and \textit{cyclist}. Boxes are drawn on every frame and linked across time.
To our best effort, the same physical instance keeps a persistent track ID throughout the clip. 
These labels directly support perception tasks such as object detection and multi-object tracking. 
In addition, we provide instance polygon masks for five infrastructure categories: \textit{ego\_lane}, \textit{other\_lane}, \textit{sidewalk}, \textit{curb}, and \textit{hard\_object}. Hence, these annotations enable tasks such as instance segmentation and lane–structure analysis.
We report annotation summary such as histograms of instance counts, track lengths, and mask areas in Supp.~\ref{sec:data_sta}.

\section{Experiment}
\label{sec:exp}

\begin{table}[t]\centering\small
\setlength{\tabcolsep}{5pt}
\resizebox{\linewidth}{!}{
\begin{tabular}{l|cc|cc|cc}
\toprule
\textbf{Metric [range]} & \multicolumn{2}{c|}{\textbf{Pika}~\cite{pika}} & \multicolumn{2}{c|}{\textbf{Sora}~\cite{sora2024}} & \multicolumn{2}{c}{\textbf{Veo\,3}~\cite{Veo3}}\\
& Raw & filtered & Raw & filtered & Raw & filtered \\
\midrule
SimpleVQA~\cite{SimpleVQA} [0,5]         & 3.5851 & \textbf{3.9214} & 3.7758 & \textbf{3.9532} & 3.7126 & \textbf{3.9831} \\
GSTVQA~\cite{GSTVQA} [0,1]               & 0.2172 & \textbf{0.2391} & 0.2679 & \textbf{0.2988} & 0.2182 & \textbf{0.2558} \\
LightVQA+~\cite{Light_VQA_plus} [0,100]  & 73.593 & \textbf{74.544} & 64.693 & \textbf{69.189} & 64.490 & \textbf{68.682} \\
DSGScore \cite{DSG}  [0, 1]              & 0.4271 & \textbf{0.4552} & 0.4511 & \textbf{0.5017} & 0.4183 & \textbf{0.4427} \\
PickScore \cite{PickScore}  [0, 1]       & 0.2235 & \textbf{0.2591} & 0.2114 & \textbf{0.2345} & \textbf{0.2577} &0.2539  \\
VIEScore \cite{ku2024viescore}  [0, 10]  & 5.2156 & \textbf{5.2218} & 5.3102 & \textbf{5.6920} & \textbf{5.9271} &5.8751  \\
CLIPTemp~\cite{liu2023evalcrafter} [0,1]  &0.8932 & \textbf{0.9127} &\textbf{0.9457} & 0.9352  & 0.9369 & \textbf{0.9693} \\
VBench~\cite{huang_vbench_2023} [0,1]     &0.8044 & \textbf{0.8965} & 0.8165 & \textbf{0.8437} & \textbf{0.8672} & 0.8564 \\
VideoPhy~\cite{bansal2024videophy} [0,1]  &0.1678 & \textbf{0.1723} & 0.1547 & \textbf{0.1969} & 0.1863 & \textbf{0.2217} \\
VideoScore~\cite{he2024videoscore} [0,4] & 2.6824 & \textbf{2.9582} & 2.9376 & \textbf{3.1783} & 2.8174 & \textbf{3.0706} \\
ADGVE [0,1]                              & 0.1834 & \textbf{0.3627} & 0.1930 & \textbf{0.3768} & 0.2018 & \textbf{0.4129} \\
\bottomrule
\end{tabular}
}
\caption{External VQA/AIGVE metrics on 50 generated 5-second clips per model before and after ADGVE filtering ($S_{\text{ADGVE}}>0.2$), at matched coverage. Across Pika, Sora, and Veo~3, ADGVE filtering removes low-quality clips and improves most distortion, temporal, and physics-aware metrics, providing evidence that ADGVE generalizes well as a model-agnostic evaluator and that ADGV-Bench, which is constructed using the same filtering policy, consists of higher-quality AI-generated driving videos. Bold numbers indicate better quality.}
\label{tab:aigve_result}
\end{table}

\newcommand{\UF}{\textcolor{gray}{U}}   
\newcommand{\FF}{\textbf{F}}            
\newcommand{\ID}{\textsc{ID}}           
\newcommand{\XD}{\textsc{Xfer}}         
\newcommand{\ap}{AP}
\newcommand{\mota}{MOTA}
\newcommand{\motp}{MOTP}
\newcommand{\miou}{mIoU}
\newcommand{\fgain}[1]{\textbf{#1}}                
\newcommand{\gain}[1]{\textcolor{gray}{\scriptsize\,(#1)}} 
\newcolumntype{L}[1]{>{\raggedright\arraybackslash}p{#1}} 
\newcolumntype{C}[1]{>{\centering\arraybackslash}p{#1}}   
\newcolumntype{A}{>{\centering\arraybackslash}p{5mm}}      

\begin{table*}[t]
\centering
\begin{tabular}{l l L{24mm} C{6mm} L{55mm}}
\toprule
\multicolumn{1}{c}{Task (metric)} &
\multicolumn{1}{c}{Model} &
\multicolumn{1}{c}{Train set: ADGV-Bench raw} &
&
\multicolumn{1}{c}{Train set: ADGV-Bench filtered} \\
\midrule
OD (AP/AP50/AP75) & DETR~\cite{DETR}   & 23.5 / 36.6 / 24.7 & $\rightarrow$ & \textbf{33.2} / \textbf{53.6} / \textbf{33.5} {\color{gray}(+9.7/+17.0/+8.8)} \\
                    & YOLOv8~\cite{YOLOv8} & 23.2 / 35.3 / 25.3 & $\rightarrow$ & \textbf{30.2} / \textbf{52.1} / \textbf{31.8} {\color{gray}(+7.0/+16.8/+6.5)} \\
\addlinespace[2pt]
MOT (MOTA/MOTP)     & ByteTrack~\cite{ByteTrack} & 48.7 / 57.2 & $\rightarrow$ & \textbf{51.9} / \textbf{62.4} {\color{gray}(+3.2/+5.2)} \\
                    & MOTR~\cite{MOTR}      & 46.3 / 55.6 & $\rightarrow$ & \textbf{68.6} / \textbf{59.3} {\color{gray}(+22.3/+3.7)} \\
\addlinespace[2pt]
IS (mIoU)      & Mask2Former~\cite{Mask2Former} & 31.1 & $\rightarrow$ & \textbf{32.9} {\color{gray}(+1.8)} \\
                    & SegFormer~\cite{SegFormer}   & 30.5 & $\rightarrow$ & \textbf{31.7} {\color{gray}(+1.2)} \\
\bottomrule
\end{tabular}
\caption{In-domain results on the ADGV-Bench test set. We train representative models for object detection (OD) , multi-object tracking (MOT), and instance segmentation (IS) on either raw ADGV-Bench or ADGVE-filtered ADGV-Bench. Training on filtered clips consistently improves AP, MOTA/MOTP, and mIoU across all tasks; gray values denote deltas relative to the raw setting next to each arrow. To isolate quality from data-volume effects, we always match the number of ADGV-Bench frames used in the raw and filtered settings across all experiments in Tables.~\ref{tab:adgve_in_domain}--\ref{tab:adgve_cross_domain_kitti360}.}
\label{tab:adgve_in_domain}
\end{table*}

\begin{table*}[!t]
\centering
\begingroup
\setlength{\tabcolsep}{4pt}                    
\renewcommand{\arraystretch}{1.05}             
\begin{tabular}{l l L{25mm} A L{27mm} A L{36mm}}
\toprule
\multicolumn{1}{c}{Task (metric)} &
\multicolumn{1}{c}{Model} &
\multicolumn{1}{c}{Train set: KITTI~\cite{KITTI}} & &
\multicolumn{1}{c}{+ADGV-Bench raw} & &
\multicolumn{1}{c}{+ADGV-Bench filtered} \\
\midrule
OD (AP/AP50/AP75) & DETR~\cite{DETR}
  & 42.0 / 62.4 / 44.2
  & $\rightarrow$
  & 39.4 / 60.6 / 42.7 {\color{gray}($-$2.6/$-$1.8/$-$1.5)}
  & $\rightarrow$
  & \textbf{44.5} / \textbf{65.2} / \textbf{46.8} {\color{gray}(+2.5/+2.8/+2.6)} \\
\bottomrule
\end{tabular}
\endgroup
\caption{Cross-domain transfer to KITTI for object detection. We compare DETR trained on KITTI only, KITTI plus raw ADGV-Bench, and KITTI plus ADGVE-filtered ADGV-Bench. Adding raw AI-generated videos hurts AP/AP50/AP75, whereas adding ADGVE-filtered videos reverses this drop and yields higher accuracy than the KITTI-only baseline; gray numbers show deltas with respect to the configuration on the left of each arrow.}
\label{tab:adgve_cross_domain_kitti}
\end{table*}

\begin{table*}[!t]
\centering
\begingroup
\setlength{\tabcolsep}{4pt}                    
\renewcommand{\arraystretch}{1.05}             
\begin{tabular}{l l L{25mm} A L{27mm} A L{36mm}}
\toprule
\multicolumn{1}{c}{Task (metric)} &
\multicolumn{1}{c}{Model} &
\multicolumn{1}{c}{Train: KITTI-360~\cite{KITTI360}} & &
\multicolumn{1}{c}{+ADGV-Bench raw} & &
\multicolumn{1}{c}{+ADGV-Bench filtered} \\
\midrule
IS (mIoU) & Mask2Former~\cite{Mask2Former}
  & 60.5
  & $\rightarrow$
  & 60.3 {\color{gray}($-$0.2)}
  & $\rightarrow$
  & \textbf{61.4} {\color{gray}(+0.9)} \\
\bottomrule
\end{tabular}
\endgroup
\caption{Cross-domain transfer to KITTI-360 for instance segmentation. Mask2Former is trained on KITTI-360 only, on KITTI-360 plus raw ADGV-Bench, and on KITTI-360 plus ADGVE-filtered ADGV-Bench. Raw ADGV-Bench slightly degrades mIoU, while ADGVE-filtered clips provide a positive gain, mirroring the trend observed for detection; gray values indicate deltas with respect to the configuration on the left of each arrow.}
\label{tab:adgve_cross_domain_kitti360}
\end{table*}

\subsection{Generated Videos Quality \& ADGVE's General Evaluation Performance Analysis}

We first examine whether ADGVE can reliably identify useful AI-generated driving videos and, in turn, whether the ADGV-Bench clips selected by ADGVE are indeed of high quality. We consider three prompt-only generators, Pika~\cite{pika}, Sora~\cite{sora2024}, and Veo~3~\cite{Veo3}, and for each model we randomly sample 50 prompts and generate one 5-second driving clip per prompt. For every clip we compute an ADGVE score and retain only those with $S_{\text{ADGVE}}>0.2$. We then evaluate both the raw and the ADGVE-filtered sets with a diverse collection of existing VQA and AIGVE metrics.

As shown in Tab.~\ref{tab:aigve_result}, ADGVE filtering consistently improves distortion and temporal/physics-focused metrics across all three generators (e.g., SimpleVQA~\cite{SimpleVQA}, GSTVQA~\cite{GSTVQA}, CLIPTemp~\cite{liu2023evalcrafter}, VideoPhy~\cite{bansal2024videophy}, and VideoScore~\cite{he2024videoscore}), while also raising most alignment-oriented metrics such as DSGScore~\cite{DSG} and VIEScore~\cite{ku2024viescore}. A few small drops on Veo~3’s alignment-heavy scores (PickScore~\cite{PickScore}, VIEScore~\cite{ku2024viescore}, VBench~\cite{huang_vbench_2023}) indicate that these metrics occasionally favor clips that are prompt-aligned but visually unstable, whereas ADGVE prioritizes temporal coherence and physical plausibility. Importantly, the ADGVE score itself almost doubles after filtering for every model, and these gains correlate well with the improvements in independent metrics. This experiment shows that ADGVE can act as a model-agnostic filter for prompt-only AIGVs. It also shows that the ADGV-Bench videos selected by ADGVE form a higher-quality and more trustworthy subset for downstream autonomous driving evaluations.

\subsection{ADGVE Filtering for Driving Perception \& Cross-Dataset Transfer}

We now evaluate whether ADGVE’s quality scores are aligned with the needs of core driving perception tasks. First, we run in-domain experiments on ADGV-Bench by training representative models for object detection, multi-object tracking, and instance segmentation on either the raw ADGV-Bench training set or its ADGVE-filtered version, and evaluating on the same test set (Tab.~\ref{tab:adgve_in_domain}). Across all tasks and models, using ADGVE-filtered clips consistently yields higher accuracy. The gains are particularly pronounced for sequence-based tracking, where removing clips with temporal flicker or implausible motion directly stabilizes training. These results indicate that ADGVE is not merely capturing generic video “pleasantness”, but identifies artifacts that are genuinely harmful for driving-specific perception models.

Next, we study cross-dataset transfer by using ADGV-Bench as an auxiliary source of AI-generated data for real-world benchmarks (Tabs.~\ref{tab:adgve_cross_domain_kitti} and \ref{tab:adgve_cross_domain_kitti360}). When we simply mix raw ADGV-Bench with KITTI~\cite{KITTI} or KITTI-360~\cite{KITTI360}, performance slightly degrades compared to training on real data alone, despite the larger training set. This confirms a key challenge of AI-generated driving videos: without quality control, they can introduce systematic noise and harm downstream perception. In contrast, augmenting with the ADGVE-filtered subset reverses this trend and even leads to clear improvements over the real-data baselines for both detection and instance segmentation. Together, these experiments show that ADGVE provides task-relevant quality assessment for driving scenarios and that its filtering enables AI-generated videos to act as a beneficial, rather than detrimental, complement to real-world datasets. Given the low marginal cost and virtually unlimited volume of prompt-only AIGVs, the pipeline in Fig.~\ref{fig:pipeline} therefore offers a highly promising recipe for next-generation driving perception, with ADGVE providing the critical quality gate that finally makes this paradigm practically usable.

\subsection{Ablations}
\label{sec:ablations}

We conduct ablations to understand which components of ADGVE are most important and how sensitive our results are to the filtering threshold. Unless otherwise stated, all ablations are performed on the ADGV-Bench validation split, and downstream experiments reuse the same backbones as in Sec.~4 (DETR and YOLOv8 for detection, ByteTrack for MOT, and Mask2Former for instance segmentation). We report Spearman’s rank correlation coefficient (SRCC), which measures the monotonic agreement between predicted scores and human ratings (SRCC $=1$ means perfect agreement, while SRCC $=0$ corresponds to random ranking).

\begin{table}[t]
\centering
\small
\begin{tabular}{lcc}
\toprule
Variant & SRCC $\uparrow$ & Avg.\ downstream $\Delta$ $\uparrow$ \\
\midrule
Full ADGVE                                 & 0.32 & +2.9 \\
w/o Object Consistency                     & 0.26 & +1.8 \\
w/o Physical Validity                      & 0.27 & +2.0 \\
w/o Semantic Plausibility                  & 0.25 & +1.7 \\
w/o Lane Obedience                         & 0.28 & +2.1 \\
w/o VLM-guided checks                      & 0.20 & +0.8 \\
\bottomrule
\end{tabular}
\caption{Module ablations for ADGVE. SRCC is the Spearman correlation with human quality scores on ADGV-Bench. Avg.\ downstream $\Delta$ is the mean improvement (in AP/MOTA/mIoU points) when replacing raw ADGV-Bench with ADGVE-filtered data for DETR, ByteTrack, and Mask2Former.}
\label{tab:ablation_modules}
\end{table}

\paragraph{Effect of individual modules.}
We first remove one module at a time and measure (i) the SRCC between predicted quality scores and human labels on ADGV-Bench, and (ii) the average improvement in downstream tasks when training on ADGVE-filtered videos versus the raw ADGV-Bench videos. For (ii), we report the mean delta across DETR AP, ByteTrack MOTA, and Mask2Former mIoU. Results are summarized in Tab.~\ref{tab:ablation_modules}.

Removing any single module lowers both alignment with human ratings and downstream gains, confirming that the taxonomy-driven design of Sec.~3.1 is complementary. The largest drop occurs when removing the VLM-guided checks, which reduces SRCC from roughly $0.32$ to $0.20$ and shrinks the average downstream gain from about $+2.9$ to $+0.8$ points. Object Consistency and Semantic Plausibility are also important, especially for tracking and segmentation, while the Lane Obedience module mainly benefits driving-rule–sensitive metrics.

\paragraph{Threshold sensitivity.}
Then, we study the impact of the quality threshold $\tau$ used to decide whether a clip is retained. We sweep $\tau \in \{0.1, 0.2, 0.3\}$ and report both coverage (fraction of clips retained) and downstream performance. Lower thresholds (e.g., $\tau = 0.1$) preserve more data but admit many low-quality clips, leading to smaller gains over the raw baseline. Higher thresholds (e.g., $\tau = 0.3$) produce very clean subsets but reduce coverage, which slightly hurts detectors that benefit from larger sample sizes. We find that $\tau = 0.2$ provides a good trade-off between quality and coverage: it retains roughly half of the generated clips while delivering the strongest average improvements across DETR, YOLOv8, ByteTrack, and Mask2Former. We adopt $\tau = 0.2$ as our default setting in Sec.~4 and provide a more detailed threshold analysis in Supp.~\ref{subsec:exp_thres}.

\section{Conclusion}
\label{sec:conclusion}

We studied whether prompt-only AI-generated driving videos can reliably support autonomous driving perception. We introduced ADGV-Bench, a driving-focused benchmark with human quality labels and downstream tasks, and ADGVE, a driving-aware evaluator that assigns each clip a single quality score. ADGVE aligns with existing VQA/AIGVE metrics, filters out harmful samples, and consistently improves detection, tracking, and segmentation when its filtered videos augment real datasets. Overall, our results expose the risks of unfiltered AIGVs and show ADGVE to be a practical gatekeeper for safely using large-scale generated videos in driving perception pipelines.
{
    \small
    \bibliographystyle{ieeenat_fullname}
    \bibliography{main}

\begin{thebibliography}{37}
\providecommand{\natexlab}[1]{#1}
\providecommand{\url}[1]{\texttt{#1}}
\expandafter\ifx\csname urlstyle\endcsname\relax
  \providecommand{\doi}[1]{doi: #1}\else
  \providecommand{\doi}{doi: \begingroup \urlstyle{rm}\Url}\fi

\bibitem[Bansal et~al.(2024)Bansal, Lin, Xie, Zong, Yarom, Bitton, Jiang, Sun, Chang, and Grover]{bansal2024videophy}
Hritik Bansal, Zongyu Lin, Tianyi Xie, Zeshun Zong, Michal Yarom, Yonatan Bitton, Chenfanfu Jiang, Yizhou Sun, Kai-Wei Chang, and Aditya Grover.
\newblock Videophy: Evaluating physical commonsense for video generation.
\newblock \emph{arXiv preprint arXiv:2406.03520}, 2024.

\bibitem[Barten(1999)]{HVS}
Peter G.~J. Barten.
\newblock \emph{Contrast Sensitivity of the Human Eye and Its Effects on Image Quality}.
\newblock SPIE Press, 1999.

\bibitem[Bartoccioni et~al.(2025)Bartoccioni, Ramzi, Besnier, Venkataramanan, Vu, Xu, Chambon, Gidaris, Odabas, Hurych, Marlet, Boulch, Chen, Éloi Zablocki, Bursuc, Valle, and Cord]{VaViM_VaVAM}
Florent Bartoccioni, Elias Ramzi, Victor Besnier, Shashanka Venkataramanan, Tuan-Hung Vu, Yihong Xu, Loick Chambon, Spyros Gidaris, Serkan Odabas, David Hurych, Renaud Marlet, Alexandre Boulch, Mickael Chen, Éloi Zablocki, Andrei Bursuc, Eduardo Valle, and Matthieu Cord.
\newblock Vavim and vavam: Autonomous driving through video generative modeling.
\newblock \emph{arXiv preprint arXiv:2502.15672}, 2025.

\bibitem[Caesar et~al.(2020)Caesar, Bankiti, Lang, Vora, Liong, Xu, Krishnan, Pan, Baldan, and Beijbom]{nuScenes}
Holger Caesar, Varun Bankiti, Alex~H. Lang, Sourabh Vora, Venice~Erin Liong, Qiang Xu, Anush Krishnan, Yu Pan, Giancarlo Baldan, and Oscar Beijbom.
\newblock {nuScenes}: A multimodal dataset for autonomous driving.
\newblock In \emph{IEEE/CVF Conference on Computer Vision and Pattern Recognition (CVPR)}, 2020.

\bibitem[Carion et~al.(2020)Carion, Massa, Synnaeve, Usunier, Kirillov, and Zagoruyko]{DETR}
Nicolas Carion, Francisco Massa, Gabriel Synnaeve, Nicolas Usunier, Alexander Kirillov, and Sergey Zagoruyko.
\newblock End-to-end object detection with transformers.
\newblock In \emph{European Conference on Computer Vision (ECCV)}. Springer, 2020.

\bibitem[Chen et~al.(2021)Chen, Zhu, Li, Fan, and Wang]{GSTVQA}
Baoliang Chen, Lingyu Zhu, Guo Li, Hongfei Fan, and Shiqi Wang.
\newblock Learning generalized spatial-temporal deep feature representation for no-reference video quality assessment.
\newblock \emph{arXiv preprint arXiv:2012.13936}, 2021.

\bibitem[Cheng et~al.(2022)Cheng, Misra, Schwing, Kirillov, and Girdhar]{Mask2Former}
Bowen Cheng, Ishan Misra, Alexander~G. Schwing, Alexander Kirillov, and Rohit Girdhar.
\newblock Masked-attention mask transformer for universal image segmentation.
\newblock In \emph{IEEE/CVF Conference on Computer Vision and Pattern Recognition (CVPR)}, 2022.

\bibitem[Cho et~al.(2024)Cho, Hu, Garg, Anderson, Krishna, Baldridge, Bansal, Pont-Tuset, and Wang]{DSG}
Jaemin Cho, Yushi Hu, Roopal Garg, Peter Anderson, Ranjay Krishna, Jason Baldridge, Mohit Bansal, Jordi Pont-Tuset, and Su Wang.
\newblock Davidsonian scene graph: Improving reliability in fine-grained evaluation for text-to-image generation.
\newblock \emph{arXiv preprint arXiv:2310.18235}, 2024.

\bibitem[{CVAT.ai Corporation}(2023)]{cvat2023}
{CVAT.ai Corporation}.
\newblock Computer vision annotation tool (cvat), 2023.

\bibitem[Dosovitskiy et~al.(2017)Dosovitskiy, Ros, Codevilla, Lopez, and Koltun]{CARLA}
Alexey Dosovitskiy, German Ros, Felipe Codevilla, Antonio Lopez, and Vladlen Koltun.
\newblock Carla: An open urban driving simulator.
\newblock In \emph{Proceedings of the 1st Annual Conference on Robot Learning (CoRL)}, pages 1--16, 2017.

\bibitem[Fu et~al.(2024)Fu, Jain, Di, Chen, and Mo]{DriveGenVLM}
Yongjie Fu, Anmol Jain, Xuan Di, Xu Chen, and Zhaobin Mo.
\newblock Drivegenvlm: Real-world video generation for vision language model based autonomous driving.
\newblock \emph{arXiv preprint arXiv:2408.16647}, 2024.

\bibitem[Geiger et~al.(2012)Geiger, Lenz, and Urtasun]{KITTI}
Andreas Geiger, Philip Lenz, and Raquel Urtasun.
\newblock Are we ready for autonomous driving? the {KITTI} vision benchmark suite.
\newblock In \emph{2012 IEEE Conference on Computer Vision and Pattern Recognition}, pages 3354--3361, 2012.

\bibitem[Google(2025)]{Veo3}
Google.
\newblock Veo 3 (gemini api) — generative video model, 2025.
\newblock Model documentation and developer reference. Accessed 2025-11-13.

\bibitem[He et~al.(2024)He, Jiang, Zhang, Ku, Soni, Siu, Chen, Chandra, Jiang, Arulraj, Wang, Do, Ni, Lyu, Narsupalli, Fan, Lyu, Lin, and Chen]{he2024videoscore}
Xuan He, Dongfu Jiang, Ge Zhang, Max Ku, Achint Soni, Sherman Siu, Haonan Chen, Abhranil Chandra, Ziyan Jiang, Aaran Arulraj, Kai Wang, Quy~Duc Do, Yuansheng Ni, Bohan Lyu, Yaswanth Narsupalli, Rongqi Fan, Zhiheng Lyu, Yuchen Lin, and Wenhu Chen.
\newblock Videoscore: Building automatic metrics to simulate fine-grained human feedback for video generation.
\newblock In \emph{Proceedings of the Empirical Methods in Natural Language Processing (EMNLP)}, pages 2105--2123. Association for Computational Linguistics, 2024.

\bibitem[Heusel et~al.(2017)Heusel, Ramsauer, Unterthiner, Nessler, and Hochreiter]{FID}
Martin Heusel, Hubert Ramsauer, Thomas Unterthiner, Bernhard Nessler, and Sepp Hochreiter.
\newblock Gans trained by a two time-scale update rule converge to a local nash equilibrium.
\newblock In \emph{Advances in Neural Information Processing Systems (NeurIPS)}, 2017.

\bibitem[Huang et~al.(2024)Huang, He, Yu, Zhang, Si, Jiang, Zhang, Wu, Jin, Chanpaisit, Wang, Chen, Wang, Lin, Qiao, and Liu]{huang_vbench_2023}
Ziqi Huang, Yinan He, Jiashuo Yu, Fan Zhang, Chenyang Si, Yuming Jiang, Yuanhan Zhang, Tianxing Wu, Qingyang Jin, Nattapol Chanpaisit, Yaohui Wang, Xinyuan Chen, Limin Wang, Dahua Lin, Yu Qiao, and Ziwei Liu.
\newblock {VBench}: {Comprehensive} {Benchmark} {Suite} for {Video} {Generative} {Models}.
\newblock In \emph{IEEE/CVF Conference on Computer Vision and Pattern Recognition (CVPR)}, 2024.

\bibitem[Jocher et~al.(2023)Jocher, Chaurasia, Qiu, et~al.]{YOLOv8}
Glenn Jocher, Ayush Chaurasia, Jing Qiu, et~al.
\newblock Ultralytics {YOLOv8}: Real-time object detection.
\newblock \url{https://docs.ultralytics.com/models/yolov8/}, 2023.
\newblock Accessed 2025-11-13.

\bibitem[Kirstain et~al.(2023)Kirstain, Polyak, Singer, Matiana, Penna, and Levy]{PickScore}
Yuval Kirstain, Adam Polyak, Uriel Singer, Shahbuland Matiana, Joe Penna, and Omer Levy.
\newblock Pick-a-pic: An open dataset of user preferences for text-to-image generation.
\newblock In \emph{Neural Information Processing Systems (NeurIPS)}, 2023.

\bibitem[Kong et~al.(2024)Kong, Tian, Zhang, Min, Dai, Zhou, Xiong, Li, Wu, Zhang, Wu, Lin, Yuan, Long, Wang, Wang, Li, Huang, Yang, Tan, Wang, Song, Bai, Wu, Xue, Wang, Wang, Liu, Li, Li, Wang, Yu, Deng, Li, Chen, Cui, Peng, Yu, He, Xu, Zhou, Xu, Tao, Lu, Liu, Zhou, Wang, Yang, Wang, Liu, Jiang, and Zhong]{HunyuanVideo}
Weijie Kong, Qi Tian, Zijian Zhang, Rox Min, Zuozhuo Dai, Jin Zhou, Jiangfeng Xiong, Xin Li, Bo Wu, Jianwei Zhang, Kathrina Wu, Qin Lin, Junkun Yuan, Yanxin Long, Aladdin Wang, Andong Wang, Changlin Li, Duojun Huang, Fang Yang, Hao Tan, Hongmei Wang, Jacob Song, Jiawang Bai, Jianbing Wu, Jinbao Xue, Joey Wang, Kai Wang, Mengyang Liu, Pengyu Li, Shuai Li, Weiyan Wang, Wenqing Yu, Xinchi Deng, Yang Li, Yi Chen, Yutao Cui, Yuanbo Peng, Zhentao Yu, Zhiyu He, Zhiyong Xu, Zixiang Zhou, Zunnan Xu, Yangyu Tao, Qinglin Lu, Songtao Liu, Dax Zhou, Hongfa Wang, Yong Yang, Di Wang, Yuhong Liu, Jie Jiang, and Caesar Zhong.
\newblock Hunyuanvideo: A systematic framework for large video generative models.
\newblock \emph{arXiv preprint arXiv:2412.03603}, 2024.

\bibitem[Ku et~al.(2024)Ku, Jiang, Wei, Yue, and Chen]{ku2024viescore}
Max Ku, Dongfu Jiang, Cong Wei, Xiang Yue, and Wenhu Chen.
\newblock Viescore: Towards explainable metrics for conditional image synthesis evaluation.
\newblock In \emph{Proceedings of the 62nd Annual Meeting of the Association for Computational Linguistics}. Association for Computational Linguistics, 2024.

\bibitem[Labs(2025)]{pika}
Pika Labs.
\newblock Pika ai video generator.
\newblock \url{https://pika.art}, 2025.
\newblock Accessed: 2025-05-22.

\bibitem[Li et~al.(2023)Li, Peng, Xue, Zhang, and Zhou]{MetaDrive}
Quanyi Li, Zhenghao Peng, Zhenghai Xue, Qihang Zhang, and Bolei Zhou.
\newblock Metadrive: Composing diverse driving scenarios for generalizable reinforcement learning.
\newblock \emph{IEEE Transactions on Pattern Analysis and Machine Intelligence}, 45\penalty0 (10):\penalty0 3461--3475, 2023.

\bibitem[Liao et~al.(2021)Liao, Xie, and Geiger]{KITTI360}
Yiyi Liao, Jun Xie, and Andreas Geiger.
\newblock Kitti-360: A novel dataset and benchmarks for urban scene understanding in 2d and 3d.
\newblock \emph{arXiv preprint arXiv:2109.13410}, 2021.

\bibitem[Liu et~al.(2024{\natexlab{a}})Liu, Xiang, Li, Wang, Li, Liu, Zhang, Ye, and Zhang]{liu2024surveyaigve}
Xiao Liu, Xinhao Xiang, Zizhong Li, Yongheng Wang, Zhuoheng Li, Zhuosheng Liu, Weidi Zhang, Weiqi Ye, and Jiawei Zhang.
\newblock A survey of ai-generated video evaluation.
\newblock \emph{arXiv preprint arXiv:2410.19884}, 2024{\natexlab{a}}.

\bibitem[Liu et~al.(2024{\natexlab{b}})Liu, Cun, Liu, Wang, Zhang, Chen, Liu, Zeng, Chan, and Shan]{liu2023evalcrafter}
Yaofang Liu, Xiaodong Cun, Xuebo Liu, Xintao Wang, Yong Zhang, Haoxin Chen, Yang Liu, Tieyong Zeng, Raymond Chan, and Ying Shan.
\newblock Evalcrafter: Benchmarking and evaluating large video generation models.
\newblock In \emph{IEEE/CVF Conference on Computer Vision and Pattern Recognition (CVPR)}, 2024{\natexlab{b}}.

\bibitem[OpenAI(2024)]{sora2024}
OpenAI.
\newblock Sora.
\newblock \url{https://openai.com/index/sora/}, 2024.

\bibitem[Sun et~al.(2020)Sun, Kretzschmar, Dotiwalla, Chouard, Patnaik, Tsui, Guo, Zhou, Chai, Caine, Vasudevan, Han, Ngiam, Zhao, Timofeev, Ettinger, Krivokon, Gao, Joshi, Zhao, Cheng, Zhang, Shlens, Chen, and Anguelov]{WOD}
Pei Sun, Henrik Kretzschmar, Xerxes Dotiwalla, Aurelien Chouard, Vijaysai Patnaik, Paul Tsui, James Guo, Yin Zhou, Yuning Chai, Benjamin Caine, Vijay Vasudevan, Wei Han, Jiquan Ngiam, Hang Zhao, Aleksei Timofeev, Scott Ettinger, Maxim Krivokon, Amy Gao, Aditya Joshi, Sheng Zhao, Shuyang Cheng, Yu Zhang, Jonathon Shlens, Zhifeng Chen, and Dragomir Anguelov.
\newblock Scalability in perception for autonomous driving: Waymo open dataset, 2020.

\bibitem[Sun et~al.(2022)Sun, Min, Lu, and Zhai]{SimpleVQA}
Wei Sun, Xiongkuo Min, Wei Lu, and Guangtao Zhai.
\newblock A deep learning based no-reference quality assessment model for ugc videos.
\newblock In \emph{Proceedings of the 30th ACM International Conference on Multimedia}, page 856–865. ACM, 2022.

\bibitem[Unterthiner et~al.(2019)Unterthiner, van Steenkiste, Kurach, Marinier, Michalski, and Gelly]{FVD}
Thomas Unterthiner, Sjoerd van Steenkiste, Karol Kurach, Raphael Marinier, Marcin Michalski, and Sylvain Gelly.
\newblock Towards accurate generative models of video: A new metric \& challenges.
\newblock \emph{arXiv preprint arXiv:1812.01717}, 2019.

\bibitem[Wang et~al.(2024)Wang, Bai, Tan, Wang, Fan, Bai, Chen, Liu, Wang, Ge, Fan, Dang, Du, Ren, Men, Zhou, Zhou, and Lin]{qwen2vl}
Peng Wang, Shuai Bai, Sinan Tan, Shijie Wang, Zhihao Fan, Jinze Bai, Keqin Chen, Xuejing Liu, Jialin Wang, Wenbin Ge, Yang Fan, Kai Dang, Mengfei Du, Xuancheng Ren, Rui Men, Chang Zhou, Jingren Zhou, and Junyang Lin.
\newblock Qwen2-vl: Enhancing vision-language model's perception of the world at any resolution.
\newblock \emph{arXiv preprint arXiv:2409.12191}, 2024.

\bibitem[Wen et~al.(2024)Wen, Zhao, Liu, Huang, Jia, Wang, Zhang, Wang, Sun, and Zhang]{Panacea+}
Yuqing Wen, Yucheng Zhao, Yingfei Liu, Binyuan Huang, Fan Jia, Yanhui Wang, Chi Zhang, Tiancai Wang, Xiaoyan Sun, and Xiangyu Zhang.
\newblock Panacea+: Panoramic and controllable video generation for autonomous driving.
\newblock \emph{arXiv preprint arXiv:2408.07605}, 2024.

\bibitem[Wilson et~al.(2024)Wilson, Qi, Agarwal, Lambert, Singh, Khandelwal, Pan, Kumar, Hartnett, Pontes, Ramanan, Carr, and Hays]{Argoverse2}
Benjamin Wilson, William Qi, Tanmay Agarwal, John Lambert, Jagjeet Singh, Siddhesh Khandelwal, Bowen Pan, Ratnesh Kumar, Andrew Hartnett, Jhony~Kaesemodel Pontes, Deva Ramanan, Peter Carr, and James Hays.
\newblock Argoverse 2: Next generation datasets for self-driving perception and forecasting.
\newblock In \emph{Neural Information Processing Systems (NeurIPS)}, 2024.

\bibitem[Xiang et~al.(2025)Xiang, Liu, Li, Liu, and Zhang]{AIGVE_Tool}
Xinhao Xiang, Xiao Liu, Zizhong Li, Zhuosheng Liu, and Jiawei Zhang.
\newblock Aigve-tool: Ai-generated video evaluation toolkit with multifaceted benchmark.
\newblock \emph{arXiv preprint arXiv:2503.14064}, 2025.

\bibitem[Xie et~al.(2021)Xie, Wang, Yu, Anandkumar, Alvarez, and Luo]{SegFormer}
Enze Xie, Wenhai Wang, Zhiding Yu, Anima Anandkumar, Jose~M. Alvarez, and Ping Luo.
\newblock Segformer: Simple and efficient design for semantic segmentation with transformers.
\newblock In \emph{Advances in Neural Information Processing Systems (NeurIPS)}, 2021.

\bibitem[Zeng et~al.(2022)Zeng, Dong, Liang, Tian, Wang, and Shen]{MOTR}
Fanqing Zeng, Bin Dong, Tingting Liang, Zhi Tian, Xinlong Wang, and Chunhua Shen.
\newblock Motr: End-to-end multiple-object tracking with transformer.
\newblock In \emph{European Conference on Computer Vision (ECCV)}. Springer, 2022.

\bibitem[Zhang et~al.(2022)Zhang, Sun, Song, Liu, Liu, Luo, Qi, and Lu]{ByteTrack}
Yifu Zhang, Peize Sun, Yibing Song, Wufei Liu, Wei Liu, Ping Luo, Donglian Qi, and Huchuan Lu.
\newblock Bytetrack: Multi-object tracking by associating every detection box.
\newblock In \emph{European Conference on Computer Vision (ECCV)}. Springer, 2022.

\bibitem[Zhou et~al.(2024)Zhou, Liu, Dong, Kou, Gao, Zhang, Li, Wu, and Zhai]{Light_VQA_plus}
Xunchu Zhou, Xiaohong Liu, Yunlong Dong, Tengchuan Kou, Yixuan Gao, Zicheng Zhang, Chunyi Li, Haoning Wu, and Guangtao Zhai.
\newblock Light-vqa+: A video quality assessment model for exposure correction with vision-language guidance.
\newblock \emph{arXiv preprint arXiv:2405.03333}, 2024.

\end{thebibliography}
}

\clearpage
\setcounter{page}{1}
\setcounter{table}{8}
\setcounter{figure}{4}
\setcounter{section}{0}
\renewcommand\thesection{\Alph{section}}
\maketitlesupplementary

\section{Additional Failure Cases of Current AI-Generated Driving Videos}
\label{sec:challenges_more}

\begin{figure*}[t]
  \centering
  \includegraphics[width=\linewidth]{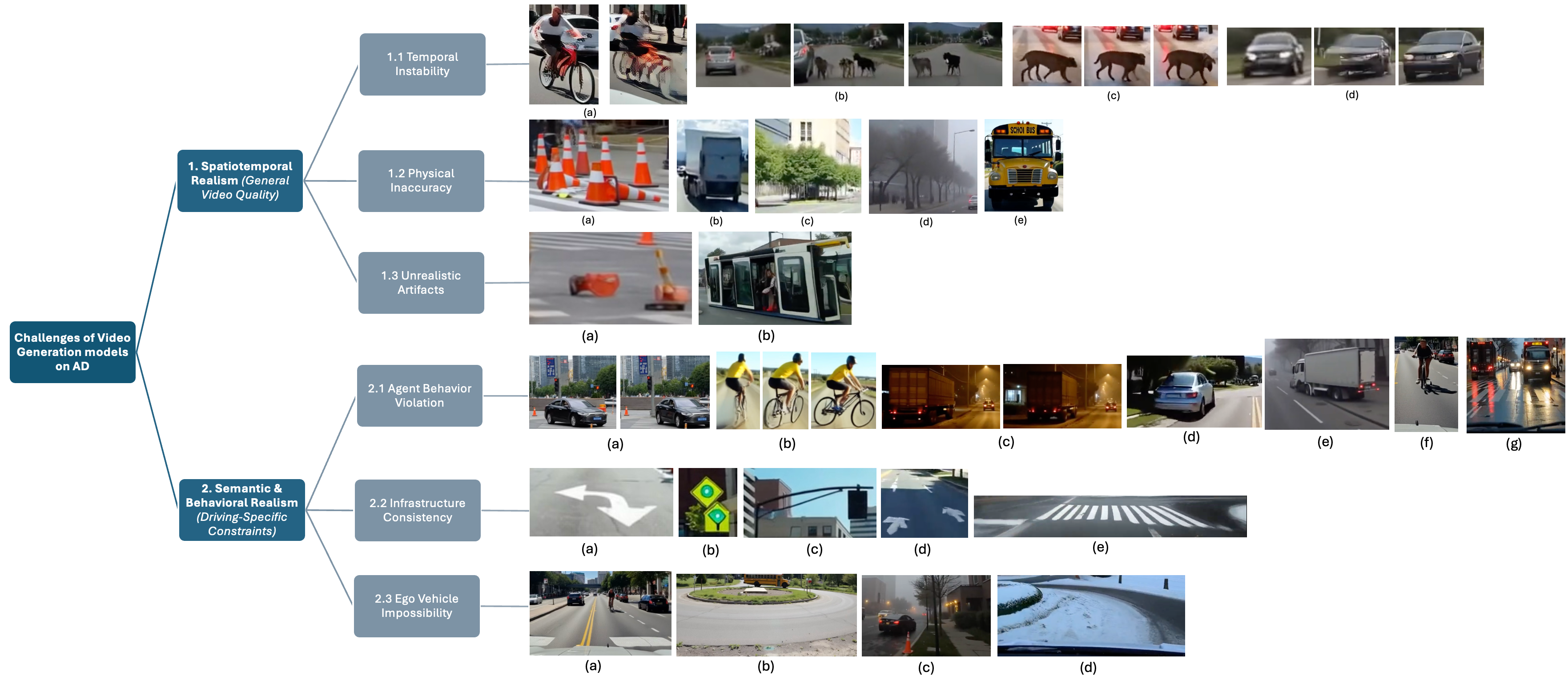}
  \caption{Additional example failure cases under challenges of currently AI-generated driving videos. 
  We briefly describe the violation for each case.
  \textbf{1.1 Temporal Instability}:
  (a) cyclist appearance flicker across frames;
  (b) cars changing shape between frames;
  (c) animal identity drift;
  (d) car front lights shifting positions.
  \textbf{1.2 Physical Inaccuracy}:
  (a) melting traffic cones;
  (b) deformed delivery truck;
  (c) trees with impossible geometry;
  (d) unrealistic fog transitions;
  (e) bus with distorted body.
  \textbf{1.3 Unrealistic Artifacts}:
  (a) floating cone base;
  (b) duplicated tram cabin.
  \textbf{2.1 Agent Behavior Violation}:
  (a) car drifting out of lane without turn;
  (b) cyclist pedaling sideways;
  (c) truck making impossible sideways motion;
  (d) ego-followed car teleporting;
  (e) truck colliding with static object;
  (f) cyclist riding in the middle of the lane;
  (g) ambulance driving against lane markings.
  \textbf{2.2 Infrastructure Consistency}:
  (a) inconsistent lane arrows;
  (b) traffic light with impossible color pattern;
  (c) misaligned traffic sign;
  (d) lane lines with broken connectivity;
  (e) zebra crossing with distorted perspective.
  \textbf{2.3 Ego Vehicle Impossibility}:
  (a) ego lane bending inconsistently over time;
  (b) ego car driving on roundabout island;
  (c) ego car parked in impossible location;
  (d) ego camera drifting off road edge.
  The relative number of examples shown under each challenge roughly reflects how frequently that failure type appears in current AI-generated driving videos. Please enlarge the figure for the best viewing experience.}
  \label{fig:challenges_more}
\end{figure*}

In this section, we extend Fig. 2 in the main paper to show additional visual failure cases for each challenge type, shown in Fig.~\ref{fig:challenges_more}. 

We systematically investigated these failure cases by applying our generated prompts to generate a large pool of AI driving clips. In total, we collected over 500 videos (more than 60,000 frames) from several popular video generators, such as Pika~\cite{pika}, Sora~\cite{sora2024}, Veo 3~\cite{Veo3}, Hunyuan~\cite{HunyuanVideo}, \textit{etc.}, covering diverse environments, weather conditions, and traffic layouts. 
We then manually screened the low-quality samples, grouped them, and finally proposed our taxonomy, which included six types of challenge.
This taxonomy focuses on cases that clearly violate either basic physical plausibility or driving rules.

Organized under this taxonomy, Fig.~\ref{fig:challenges_more} clearly illustrates how current AI-generated driving videos can violate both general video quality and driving-specific constraints.
These issues directly undermine core perception functions.
For example, object detection and tracking tasks struggle when appearance and geometry are unstable.
In addition while navigation, mapping, and planning tasks become unreliable when lanes, traffic lights, or other agents do not obey basic traffic rules. 
As a result, naively using raw AI-generated videos as training or evaluation data for autonomous driving can introduce harmful biases and degrade performance rather than provide useful augmentation. 
These diverse failure modes highlight the need for a dedicated video evaluator that can go beyond generic “visual quality” scoring and explicitly account for driving-aware semantics and constraints, motivating our ADGVE framework.

\section{Video-Generation Instructions: Vocabulary, Priors, Coverage in ADGV-Bench}
\label{sec:instructions}

\begin{table*}[t]
\small
\setlength{\tabcolsep}{4pt}
\centering
\begin{tabular}{lcp{13cm}}
\toprule
\textbf{Factor dimension} & \textbf{Prior} & \textbf{Vocabulary} \\
\midrule
\multirow{3}{*}{Environment ($E$)}
  & high   & urban, residential, highway \\
  & medium & suburban, rural \\
  & low    & mountain, tunnel, bridge, roundabout, parking lot \\
\midrule
\multirow{3}{*}{Weather ($W$)}
  & high   & sunny, overcast \\
  & medium & rainy, night \\
  & low    & snowy, foggy, heavy rain/snow \\
\midrule
\multirow{3}{*}{Ego behavior ($B$)}
  & high   & cruising straight, lane keeping \\
  & medium & braking, overtaking, merging, navigating traffic,
             stop-and-go traffic, approaching traffic light \\
  & low    & U-turn, emergency braking, exiting highway \\
\midrule
\multirow{3}{*}{Scene dynamics ($D$)}
  & high   & sparse traffic, heavy traffic, parked cars \\
  & medium & construction obstacle (roadwork cones, temporary barriers), 
             visual obstruction (glare, strong headlights, smoke),
             pedestrians at crosswalk, bus stopping, truck blocking lane \\
  & low    & emergency vehicle, jaywalker, cyclist in ego lane,
             animal presence, small obstacle on road \\
\bottomrule
\end{tabular}
\caption{Factor vocabularies and relative sampling priors used to construct
ADGV\mbox{-}Bench. Each factor value used in
Table~\ref{tab:supp_prompts_examples} is drawn from this vocabulary. Values
with \textbf{high}, \textbf{medium}, or \textbf{low} prior are sampled more or
less frequently to balance everyday and challenging driving scenarios.}
\label{tab:supp_factor_vocab}
\end{table*}

\begin{table*}[t]
\small
\centering
\setlength{\tabcolsep}{4pt}
\begin{tabular}{p{1.9cm}cccp{2cm}p{6.1cm}}
\toprule
ID & $E$ (Env.) & $W$ (Weather) & $B$ (Ego beh.) & $D$ (Scene dyn.) &
Generated Video Instruction \\
\midrule
ADGVE-Combo-001 & Urban       & Sunny    & Braking            & Construction obstacle (roadwork cones) &
In the driver's front-camera view of a sunny urban downtown, the ego car brakes for construction obstacles and bright orange roadwork cones blocking the lane. \\
ADGVE-Combo-011 & Urban       & Foggy    & Overtaking         & Construction obstacle (temporary barriers) &
From the front camera in a foggy city morning, the ego car overtakes a slow vehicle while squeezing past a row of temporary concrete barriers closing one lane. \\
ADGVE-Combo-021 & Highway     & Snowy    & Navigating traffic & Heavy traffic &
On a snowy evening highway scene, the ego vehicle carefully navigates heavy traffic with trucks and cars filling multiple lanes at different speeds. \\
ADGVE-Combo-041 & Suburban    & Night    & Overtaking         & Visual obstruction (strong headlights) &
At suburban dusk, the ego car overtakes a slow vehicle while strong oncoming headlights and street lights create intense visual obstruction ahead. \\
ADGVE-Combo-061 & Residential & Rainy    & Braking            & Heavy traffic &
In a rainy residential school zone, the ego vehicle brakes for a pedestrian while nearby cars create slow-moving heavy traffic. \\
ADGVE-Combo-081 & Suburban    & Sunny    & Merging            & Emergency vehicle &
On a sunny suburban expressway on-ramp, the ego car merges into traffic as an emergency vehicle with flashing lights approaches fast from behind. \\
ADGVE-Combo-091 & Highway     & Rainy    & Lane keeping       & Sparse traffic &
On a rainy highway afternoon, the ego car keeps its lane at steady speed while only a few distant vehicles appear ahead in sparse traffic. \\
ADGVE-Combo-101 & Urban       & Night    & Cruising straight  & Pedestrians at crosswalk &
At night on an urban main street, the ego car cruises straight and slows gently as groups of pedestrians begin crossing at a signalized crosswalk. \\
ADGVE-Free-001  & Residential & Overcast & Overtaking         & Animal presence &
On an overcast residential street, the ego vehicle overtakes a parked car while a dog crosses near the curb, briefly distracting the driver. \\
ADGVE-Free-021  & Rural       & Snowy    & Merging            & Emergency vehicle &
In a snowy rural setting, the ego car merges onto a narrow road while an emergency vehicle speeds past with limited visibility. \\
ADGVE-Free-041  & Mountain    & Foggy    & Navigating traffic & Emergency vehicle &
On a foggy mountain road, the ego vehicle carefully navigates light traffic as an emergency vehicle with siren on suddenly appears from behind and passes. \\
ADGVE-Free-061  & Rural       & Snowy    & Lane keeping       & Jaywalker &
Along a snowy rural main street, the ego car keeps its lane while a lone pedestrian jaywalks across the road far from any marked crosswalk. \\
ADGVE-Free-081  & Urban       & Heavy rain/snow & Stop-and-go traffic & Cyclist in ego lane &
In heavy mixed rain and snow on a congested urban street, the ego car moves in stop-and-go traffic while a cyclist briefly rides in the ego lane ahead. \\
\bottomrule
\end{tabular}
\caption{Representative generated video instructions in ADGV\mbox{-}Bench and
their factor assignments. Each row shows one concrete instruction together
with its Environment ($E$), Weather ($W$), Ego behavior ($B$), and Scene
dynamics ($D$) values, all drawn from the vocabulary in
Table~\ref{tab:supp_factor_vocab}. We include both structured prompts
(ADGVE\mbox{-}Combo) and free-style prompts (ADGVE\mbox{-}Free).}
\label{tab:supp_prompts_examples}
\end{table*}

This section complements the `Prompt Instruction Generation' part in Sec.~3.3 of the main paper by providing the full factor vocabularies, sampling priors, and additional video-generation instruction examples used to construct ADGV-Bench and to serve as text prompts for the video generators.
We keep the same factorization of video-generation instructions into
\emph{Environment} ($E$), \emph{Weather} ($W$), \emph{Ego behavior} ($B$), and
\emph{Scene dynamics} ($D$), and the same two-generation strategies
(structured templates and free-style prompts); here we focus only on the extra
details omitted from the main paper.
Each instruction in this section is instantiated as a natural-language prompt and sent to all prompt-only video generators considered in ADGV-Bench, ensuring that differences across models come from generation behavior rather than from the underlying instruction set.

\paragraph{Factor vocabularies and sampling priors.}
For each factor we define a discrete vocabulary of candidate phrases, shown in Table~\ref{tab:supp_factor_vocab}.  
\textbf{\textit{Environments $E$}} cover common road layouts(urban/residential streets, multi-lane city avenues, highways, rural and mountain roads, tunnels, roundabouts, parking lots, \emph{etc.}); 
\textbf{\textit{Weather $W$}}spans clear and overcast day, sunset/dawn, night, rain of different intensities, snow, fog, and wet-road conditions; 
\textbf{\textit{Ego behaviors $B$}} include cruising, lane keeping, lane changes, merges, exits, left/right turns, stop-and-go traffic, approaching or leaving intersections and traffic lights, and occasional emergency maneuvers; 
\textbf{\textit{Scene dynamics $D$}} include static infrastructure (parked cars, work zones, cones, barriers) as well as dynamic agents (oncoming traffic, cyclists, groups of pedestrians, jaywalkers, buses/trucks, emergency vehicles, vehicles cutting in, small obstacles on the road, \emph{etc.}).  

We assign each phrase a sampling prior so that everyday scenarios (e.g., clear urban driving with moderate traffic) occur more frequently, while allocating non-trivial probability mass to challenging conditions such as night, rain/snow, unusual maneuvers, or abnormal agent behavior. 
These priors are shared across all generators so that differences in dataset coverage come from the models rather than from the prompt distribution itself.

\paragraph{Factor-wise design considerations.}
These factor dimensions are chosen to align with how downstream driving tasks parse the scene.  
The \textit{environment factor $E$} controls large-scale layout and viewpoint and is designed to cover both typical navigation (straight urban roads, intersections, highways) and topology that is notoriously difficult for perception (roundabouts, tunnels, bridges).  
The \textit{weather factor $W$} targets conditions that significantly change image statistics and visibility, which is crucial for robustness studies.  
\textit{Ego behavior $B$} focuses on the motion of the ego vehicle, explicitly including maneuvers that induce strong camera motion (lane changes, turns, merges) so that our prompts do not collapse to ``driving straight''.  
Finally, \textit{Scene dynamics $D$} captures the presence and interaction of other agents and infrastructure. 

We intentionally up-weight cases that tend to create safety-critical situations in real traffic (e.g., pedestrians in the ego lane, vehicles cutting in, emergency vehicles, roadwork cones).  
When sampling video-generation instructions (and thus prompts to the generators), we also reject combinations that are physically or legally
contradictory (e.g., ``clear night sun'') and lightly rebalance rare but
important settings so that each factor value appears in a sufficient number of
instructions.

\paragraph{Additional prompt examples.}
Table~\ref{tab:supp_prompts_examples} lists a set of representative video-generation instructions (i.e., text prompts) together with their $(E, W, B, D)$ assignments. For clarity, we show both structured and free-style prompts. 
For example, a structured prompt may read:
\emph{``In the driver's front-camera view of a multi-lane urban avenue on a rainy night, the ego car changes lanes while roadwork cones and a slow delivery truck appear ahead.''}  A corresponding free-style prompt using the same factors dimension might say: \emph{``At night in the city, rain streaks across the windshield as the car weaves around orange construction cones and a lumbering delivery truck in the next lane.''}  
Across the full bank of 150 prompts, we verify that the empirical distribution over $(E,W,B,D)$ combinations covers a wide range of environments, weather conditions, ego maneuvers, and dynamic events, rather than collapsing to a few easy ``sunny straight-road'' cases. 
This ensures that ADGV-Bench probes both typical and challenging driving
situations under a controlled yet diverse prompt design.

\section{ADGV-Bench Dataset Statistics and Annotation Protocol}
\label{sec:data_sta}

\begin{table}[t]
\small
\centering
\setlength{\tabcolsep}{4pt}
\begin{tabular}{lcc}
\toprule
Actor category & \# tracks & \# bounding boxes \\
\midrule
Vehicle    & 27361 & 228360 \\
Pedestrian & 15383 & 103773 \\
Cyclist    & 4742 & 21642 \\
\midrule
\textbf{Total} & \textbf{47486} & \textbf{353775} \\
\bottomrule
\end{tabular}
\caption{Bounding-box statistics for traffic actors in ADGV\mbox{-}Bench.
Each actor instance is annotated with a 2D bounding box on every frame where
it is visible and assigned a persistent track ID across time.}
\label{tab:supp_bbox_stats}
\end{table}

\begin{table}[t]
\small
\centering
\setlength{\tabcolsep}{4pt}
\begin{tabular}{lcc}
\toprule
Mask category & \# instances (polygons) \\
\midrule
Ego\_lane    & 12832  \\
Other\_lane  & 11273 \\
Sidewalk     & 952 \\
Curb         & 23273  \\
Hard\_object & 44379  \\
\midrule
\textbf{Total} & \textbf{92709} \\
\bottomrule
\end{tabular}
\caption{Instance mask statistics for infrastructure categories in
ADGV\mbox{-}Bench. Each instance is annotated with a polygon mask on every
frame where it is visible.}
\label{tab:supp_mask_stats}
\end{table}

\begin{figure*}[t]
\centering
\includegraphics[width=\linewidth]{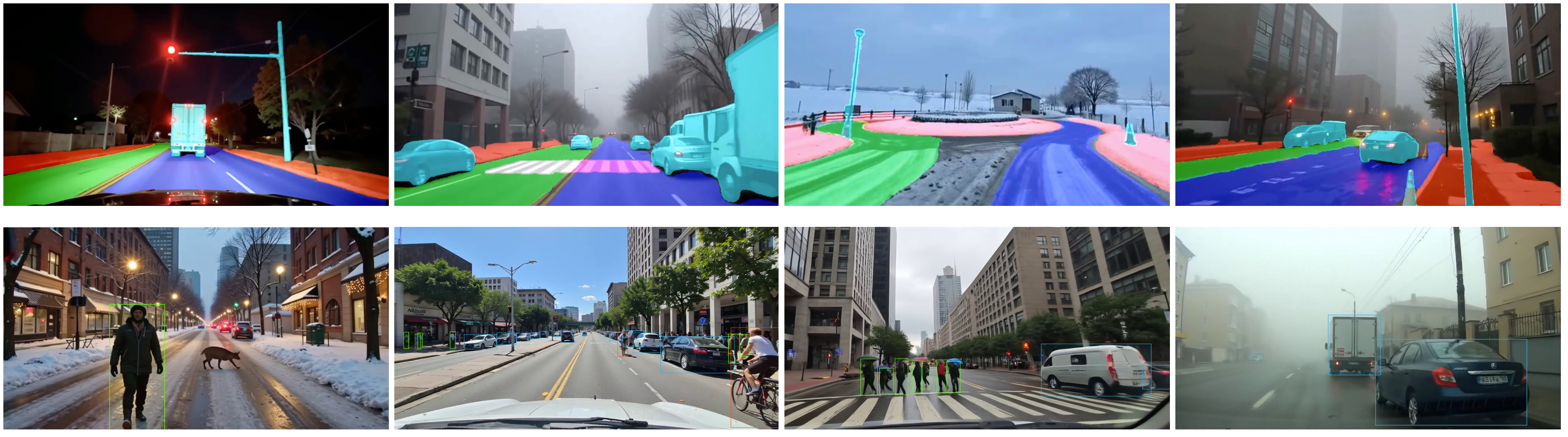}
\caption{Example annotations in ADGV\mbox{-}Bench. \textbf{Top row:} polygon
instance masks for ego\_lane, other\_lane, sidewalk, curb, and hard\_object
across diverse driving scenes. \textbf{Bottom row:} 2D bounding boxes for vehicles, pedestrians, and cyclists in various traffic conditions.
Different colors denote different semantic categories.}
\label{fig:supp_annotation_interface}
\end{figure*}

This section complements the `Video Annotation' part in Sec.~3.3 of the main paper by providing additional statistics and details of our annotation protocol. 
Following KITTI~\cite{KITTI}, for each generated video we annotate 2D object bounding boxes and tracks for three traffic categories (\textit{vehicle, pedestrian, cyclist}), and polygon instance masks for five infrastructure categories (\textit{ego\_lane, other\_lane, sidewalk, curb, hard\_object}). 
These labels could support down-streaming perception tasks such as object detection, multi-object tracking, instance segmentation, and lane–structure analysis.

\paragraph{Overall statistics.}
ADGV\mbox{-}Bench currently contains 90 videos and 10890 frames in total. Across all frames we annotate 353775 object bounding boxes grouped into 47486 tracks, and 92709 polygon masks for the five infrastructure categories. 
Tables~\ref{tab:supp_bbox_stats} and \ref{tab:supp_mask_stats} break these numbers down by category. 
These statistics confirm that \textit{vehicles} dominate the traffic participant label, while \textit{pedestrians} and \textit{cyclists} still appear in sufficient quantities to support evaluation of rare but safety-critical interactions. 
For masks, \textit{ego\_lane} and \textit{other\_lane} cover most pixels, while \textit{curb} and \textit{hard\_object} provide sparse but important structural cues.

\paragraph{Annotation protocol.}
We annotate all videos using a professional video-annotation tool~\cite{cvat2023} with
frame-by-frame playback and track management (see Fig.~\ref{fig:supp_annotation_interface}). Annotators follow a unified set of
guidelines:

\begin{itemize}
  \item \textbf{Object bounding boxes.}
  For \textit{vehicles}, \textit{pedestrians}, and \textit{cyclists}, we draw a tight 2D bounding box around the visible extent of the object on every frame where it appears.
  Each physical instance is assigned a unique track ID that remains constant over time. When an object is fully occluded or outside the image, the track is temporarily suspended and re-activated when it re-enters the view.
  \item \textbf{Instance segmentation masks.}
  We annotate polygons for five infrastructure categories: \textit{ego\_lane} (drivable lane where the ego vehicle is currently located), \textit{other\_lane} (adjacent or oncoming lanes), \textit{sidewalk}, \textit{curb}, and \textit{hard\_object} (e.g., traffic cones, bollards, barriers, poles). Polygons are required to follow visible boundaries as closely as possible and to remain temporally consistent across frames.
  \item \textbf{Ambiguous cases.}
  When category boundaries are ambiguous (e.g., painted islands vs.\ curb, parking spaces vs.\ lane), annotators follow a written decision list and consult example screenshots. Difficult clips are flagged and reviewed by a second annotator.
  \item \textbf{Quality control.}
  We perform a second-pass audit on a random subset of videos, checking for missing objects, tracking ID switches, polygon leakage, and label consistency across time. Clips with high error rates are sent back for correction.
\end{itemize}

\paragraph{Visualization.}
Figure~\ref{fig:supp_annotation_interface} illustrates typical annotations in
ADGV\mbox{-}Bench, including bounding boxes for traffic participants and polygon masks for
lanes and infrastructure.

\section{Implementation Details of the ADGVE Evaluator}

\subsection{Prompt Building: Full Prompt Set}
\label{subsection:full_set}


\paragraph{Prompt structure overview.}
ADGVE uses a unified prompting scheme for all visual--language queries.
Each query consists of two components: (i) a \emph{global instruction}
that defines the VLM's role and output format, and (ii) a
\emph{query-specific template} that specifies the question and a finite
set of candidate descriptions. The visual input can be a raw frame,
a rendered frame with overlays, a short clip, or an ROI crop, while
the textual input is the concatenation of the global instruction and
the instantiated query template. Detector/segmenter/tracker outputs are
used only to define ROIs and colored overlays in the rendered images;
their latent features are \emph{not} fed into the VLM.

\paragraph{Global instruction (shared).}
All VLM queries share the same textual prefix that specifies the
evaluator's role and the desired output format:
\begin{quote}
\small
\texttt{You are a careful driving-scene evaluator. For each question you will see a driving image or video clip and a candidate description of the scene. Decide whether the description is consistent with the visual evidence and state how confident you are in your choice. Reply strictly in the format:\{'answer': Yes/No, 'confidence': x.xx\}, where confidence is in range [0,1].}
\end{quote}
This instruction is prepended to every question template described
below; the question text and its candidate descriptions are provided and
grouped in the following paragraphs.

\paragraph{Candidate-wise querying and feature encoding.}
Following the spirit of LightVQA+~\cite{Light_VQA_plus}, each
query-specific template contains a single placeholder token
\texttt{[ANSWER]} inside a short sentence (e.g.,
\texttt{``This is a [ANSWER] driving image.''}). Let
$\mathcal{C} = \{c_1,\dots,c_{|\mathcal{C}|}\}$ denote the set of
answer candidates for a given query. We implement each check as a series
of Yes/No questions. For each candidate $c \in \mathcal{C}$, we instantiate one prompt by replacing \texttt{[ANSWER]} with $c$ and concatenate it with the shared instruction.
Given a fixed visual input (frame, clip, or ROI), we query Qwen2--VL~\cite{qwen2vl} with this text and parse the returned confidence ${\rm conf}(c) \in [0,1]$ for the ``Yes'' answer. 
We form a candidate-wise feature vector ${\bm p} = [{\rm conf}(c_1),\dots,{\rm conf}(c_{|\mathcal{C}|})]$ and optionally apply a simple per-query normalization (e.g., dividing by $\sum_c {\rm conf}(c)$ or a softmax) so that ${\bm p}$ behaves as a probability distribution over $\mathcal{C}$. This step does not introduce any new learnable parameters.

For each query, the resulting vector ${\bm p}$ is used as the query's feature representation. 
All such vectors from Group~A/B/C are concatenated into the frame- and clip-level features that feed the score-fusion module in Sec.~3.2 of the main paper.
We do not introduce any additional hand-crafted mapping from candidates to scalar scores here, and the only learnable mapping is the global fusion regressor shared by all queries.

\paragraph{Group A -- Frame-level quality and realism (key frames).}
We assess general visual quality on single key frames. For each video we
sample one representative key frame and feed the full-frame image to the
VLM for the following checks.

\begin{enumerate}[leftmargin=1.6em,label=\textbf{A\arabic*.}]
\item \textbf{Brightness / exposure.} \\
\textit{Target challenges:} 1.1 Temporal Instability. \\
\textit{Inputs:} full-frame key image. \\
\textit{Candidates:} \{underexposed, slightly-underexposed, well-exposed,
slightly-overexposed, overexposed\}. \\
\textit{Prompt:} \emph{``This is a [ANSWER] driving image.''}

\item \textbf{Sharpness / clarity.} \\
\textit{Target challenges:} 1.1 Temporal Instability, 1.2 Physical Inaccuracy. \\
\textit{Inputs:} full-frame key image. \\
\textit{Candidates:} \{very-blurry, slightly-blurry, clear, very-sharp\}. \\
\textit{Prompt:} \emph{``This is a [ANSWER] driving image in terms of sharpness and clarity.''}

\item \textbf{Scene completeness.} \\
\textit{Target challenges:} 1.3 Unrealistic Artifacts. \\
\textit{Inputs:} full-frame key image. \\
\textit{Candidates:} \{complete-road-scene, partly-missing-road,
heavily-corrupted\}. \\
\textit{Prompt:} \emph{``This frame shows a [ANSWER].''}

\item \textbf{Noise / cleanliness.} \\
\textit{Target challenges:} 1.3 Unrealistic Artifacts. \\
\textit{Inputs:} full-frame key image. \\
\textit{Candidates:} \{very-noisy, noisy, moderate-noise, clean\}. \\
\textit{Prompt:} \emph{``In terms of digital noise and grain, this is a [ANSWER] driving image.''}

\item \textbf{Color realism.} \\
\textit{Target challenges:} 1.2 Physical Inaccuracy, 1.3 Unrealistic Artifacts. \\
\textit{Inputs:} full-frame key image. \\
\textit{Candidates:} \{strongly-unnatural-colors, slightly-unnatural-colors,
mostly-natural-colors, very-natural-colors\}. \\
\textit{Prompt:} \emph{``Overall, the colors and tones in this driving image are [ANSWER].''}

\item \textbf{Single-frame layout plausibility.} \\
\textit{Target challenges:} 1.2 Physical Inaccuracy, 1.3 Unrealistic Artifacts. \\
\textit{Inputs:} full-frame key image. \\
\textit{Candidates:} \{geometrically-plausible, slightly-strange-layout,
impossible-or-highly-distorted\}. \\
\textit{Prompt:} \emph{``Overall, the road, buildings, and horizon in this image look [ANSWER].''}
\end{enumerate}

\paragraph{Group B -- Clip-level temporal consistency (short clips).}
We then assess temporal stability on short clips. For each video we
render a sub-clip $Y_m$ with $K$ frames (typically $K{=}8$) at a fixed
stride.

\begin{enumerate}[leftmargin=1.6em,label=\textbf{B\arabic*.}]
\item \textbf{Appearance flicker.} \\
\textit{Target challenges:} 1.1 Temporal Instability. \\
\textit{Inputs:} full sub-clip $Y_m$. \\
\textit{Candidates:} \{no-flicker, mild-flicker, strong-flicker\}. \\
\textit{Prompt:} \emph{``Across these frames, the colors and textures exhibit [ANSWER].''}

\item \textbf{Object identity stability.} \\
\textit{Target challenges:} 1.1 Temporal Instability. \\
\textit{Inputs:} cropped tracklets overlaid with boxes and IDs. \\
\textit{Candidates:} \{consistent-identities, slightly-inconsistent,
highly-inconsistent\}. \\
\textit{Prompt:} \emph{``Across this clip, the identities and
appearances of vehicles and pedestrians are [ANSWER].''}

\item \textbf{Camera motion smoothness.} \\
\textit{Target challenges:} 1.1 Temporal Instability, 1.2 Physical Inaccuracy. \\
\textit{Inputs:} full sub-clip $Y_m$. \\
\textit{Candidates:} \{very-smooth, slightly-shaky, strongly-shaky\}. \\
\textit{Prompt:} \emph{``Across this clip, the camera motion is [ANSWER].''}

\item \textbf{Parallax and depth consistency.} \\
\textit{Target challenges:} 1.2 Physical Inaccuracy. \\
\textit{Inputs:} full sub-clip $Y_m$. \\
\textit{Candidates:} \{physically-consistent, slightly-inconsistent,
clearly-inconsistent\}. \\
\textit{Prompt:} \emph{``The relative motion of near and far objects in this clip is [ANSWER] with real-world parallax.''}

\item \textbf{Ego-lane centering (trajectory smoothness).} \\
\textit{Target challenges:} 1.1 Temporal Instability, 2.3 Ego Vehicle Impossibility. \\
\textit{Inputs:} full sub-clip $Y_m$ with ego-lane mask overlay. \\
\textit{Candidates:} \{well-centered-in-lane, slightly-drifting,
frequently-out-of-lane\}. \\
\textit{Prompt:} \emph{``Relative to the highlighted ego lane, the camera trajectory is [ANSWER].''}

\item \textbf{Agent motion naturalness (vehicles and pedestrians).} \\
\textit{Target challenges:} 1.2 Physical Inaccuracy, 2.1 Agent Behavior Violation. \\
\textit{Inputs:} cropped tracklets of nearby agents with boxes and IDs. \\
\textit{Candidates:} \{natural-and-smooth, slightly-unnatural-or-jerky,
highly-unnatural-or-teleporting\}. \\
\textit{Prompt:} \emph{``Across this clip, the motions of vehicles and pedestrians are [ANSWER].''}

\item \textbf{Ego speed reasonableness.} \\
\textit{Target challenges:} 1.2 Physical Inaccuracy, 2.3 Ego Vehicle Impossibility. \\
\textit{Inputs:} full sub-clip $Y_m$ showing ego motion and context. \\
\textit{Candidates:} \{too-slow-for-scene, reasonable-for-scene,
too-fast-for-scene\}. \\
\textit{Prompt:} \emph{``Given the road type and surroundings, the ego-vehicle speed in this clip is [ANSWER].''}
\end{enumerate}

\paragraph{Group C -- Driving-rule and layout checks (ROI overlays).}
Finally, we probe driving-specific semantics using overlays on object
and infrastructure ROIs derived from masks and tracks. These queries aim
to capture violations such as driving on sidewalks or unclear lane
markings.

\begin{enumerate}[leftmargin=1.6em,label=\textbf{C\arabic*.}]
\item \textbf{Forbidden-region occupancy (vehicle).} \\
\textit{Target challenges:} 2.1 Agent Behavior Violation, 2.3 Ego Vehicle Impossibility. \\
\textit{Inputs:} vehicle ROI; overlay forbidden masks (sidewalk, opposite
lane, median) in red. \\
\textit{Candidates:} \{entirely-in-drivable-lane, partly-on-sidewalk,
in-opposite-lane, on-shoulder-or-median\}. \\
\textit{Prompt:} \emph{``Relative to the highlighted forbidden regions, this vehicle is [ANSWER].''}

\item \textbf{Forbidden-region occupancy (pedestrian/cyclist).} \\
\textit{Target challenges:} 2.1 Agent Behavior Violation. \\
\textit{Inputs:} pedestrian/cyclist ROI; overlay ego\_lane / road region
in red. \\
\textit{Candidates:} \{on-sidewalk-or-crosswalk, partly-in-ego-lane,
mostly-in-ego-lane\}. \\
\textit{Prompt:} \emph{``Relative to the vehicle lane, this person or
cyclist is [ANSWER].''}

\item \textbf{Lane / arrow clarity.} \\
\textit{Target challenges:} 2.2 Infrastructure Consistency. \\
\textit{Inputs:} lane and arrow mask ROI; overlay mask boundary. \\
\textit{Candidates:} \{clear-and-correct, faded-or-broken,
ambiguous-or-missing\}. \\
\textit{Prompt:} \emph{``The lane markings and arrows in this region are [ANSWER].''}

\item \textbf{Traffic-signal correctness (appearance).} \\
\textit{Target challenges:} 2.2 Infrastructure Consistency. \\
\textit{Inputs:} traffic light/sign ROI with overlay. \\
\textit{Candidates:} \{realistic-and-properly-placed,
distorted-or-unrealistic\}. \\
\textit{Prompt:} \emph{``The traffic light or sign in this region is [ANSWER].''}

\item \textbf{Crosswalk right-of-way.} \\
\textit{Target challenges:} 2.1 Agent Behavior Violation. \\
\textit{Inputs:} ROI covering a crosswalk region over a short clip, with pedestrians and nearby vehicles overlaid. \\
\textit{Candidates:} \{vehicles-yielding-properly, vehicles-not-yielding,
no-clear-conflict\}. \\
\textit{Prompt:} \emph{``Considering this crosswalk scene, the vehicle behavior toward pedestrians is [ANSWER].''}

\item \textbf{Solid-line respect.} \\
\textit{Target challenges:} 2.1 Agent Behavior Violation, 2.3 Ego Vehicle Impossibility. \\
\textit{Inputs:} lane-boundary ROI over time; overlay solid lane lines in red. \\
\textit{Candidates:} \{no-solid-line-crossing, brief-or-minor-crossing,
clear-or-prolonged-crossing\}. \\
\textit{Prompt:} \emph{``Relative to the highlighted solid lane boundaries, the vehicle behavior is [ANSWER].''}

\item \textbf{Stopping / parking legality.} \\
\textit{Target challenges:} 2.1 Agent Behavior Violation, 2.2 Infrastructure Consistency. \\
\textit{Inputs:} ROI around a stopped or parked vehicle with lane and roadside masks overlaid. \\
\textit{Candidates:} \{properly-parked-or-in-designated-bay,
briefly-stopped-in-lane, illegally-stopped-or-blocking-lane\}. \\
\textit{Prompt:} \emph{``Relative to the lane and roadside markings, this stopped vehicle is [ANSWER].''}
\end{enumerate}

\subsection{Lane obedience module}
\label{subsection:lane_obedience}
The lane-obedience score $s_{lane}$ combines three normalized
components computed from semantic masks $\mathcal{M}$ and tracklets $\mathcal{T}$.

\emph{(i) Lane centering.}
From the `ego\_lane' and `other\_lane' masks we extract lane centerlines
by applying skeletonization followed by polyline simplification.
For each vehicle tracklet (including the ego vehicle when available)
we sample points along its trajectory and compute the distance to the
closest lane centerline in pixels, ignoring frames where the vehicle
is fully off-road. The per-video lane-centering error
$d_{norm}$ is the average distance divided by the local lane
width. We then map this error to $[0,1]$ via
$s_{center} = \exp(-\alpha d_{norm})$ with
$\alpha = 1.0$ by default, so well-centered trajectories yield scores
close to~1 and poorly centered ones decay smoothly toward~0.

\emph{(ii) Solid-line violations.}
We classify lane boundaries as solid vs.\ dashed from the lane masks
and rasterize them as thin polylines. For every vehicle trajectory we
count the number of times its footprint crosses a solid boundary
between adjacent lanes. The violation rate is the number of violating
segments divided by the total number of valid trajectory segments.
We then define $s_{solid} = 1 - violation\_rate$,
clipped to $[0,1]$, so videos with frequent solid-line crossings
receive low scores.

\emph{(iii) Crosswalk yielding.}
From crosswalk regions (a subset of the hard\_object / road-marking
masks) we derive a binary occupancy signal: at each frame we check
whether any pedestrian track lies inside the crosswalk and whether
the ego-vehicle trajectory enters a $d$-meter neighborhood upstream
of that crosswalk (using pixel-to-meter calibration from lane width).
If the ego car enters this neighborhood while the crosswalk is
occupied and its speed is above a threshold, we mark a yielding
violation. The crosswalk score is then
$s_{cross} = 1 - violation\_freq$, again clipped to
$[0,1]$, where $violation\_freq$ is the fraction of such
ego–crosswalk encounters that are violations.

Finally, we combine the three components into a single lane-obedience
score
\begin{equation}
s_{lane}
= w_{center} s_{center}
+ w_{solid} s_{solid}
+ w_{cross} s_{cross},
\end{equation}
with fixed weights
$(w_{center}, w_{solid}, w_{cross})
= (0.4, 0.3, 0.3)$ on all splits.
Each component is computed per video and then averaged over all clips
from the same sequence. The resulting $s_{lane}$ is passed to
the fusion regressor together with VLM-derived and static/temporal
features, as in Eq.~(2) of the main paper, and also serves as an
interpretable breakdown of lane obedience in our diagnostic analysis.





\section{Extended Experiments and Sensitivity Analyses}

\subsection{Threshold Sensitivity of ADGVE}
\label{subsec:exp_thres}

\begin{table}[t]
\centering
\small
\resizebox{\linewidth}{!}{
\begin{tabular}{lcccc}
\toprule
$\tau$ & Coverage (\%) & $\Delta$OD (AP) & $\Delta$MOT (MOTA) & $\Delta$IS (mIoU) \\
\midrule
0.1 & 68.4 & +0.7 & +0.4 & +0.6 \\
0.2 & 52.1 & +2.1 & +1.2 & +1.7 \\
0.3 & 37.3 & +1.4 & +0.8 & +1.1 \\
\bottomrule
\end{tabular}
}
\caption{Effect of the ADGVE threshold $\tau$ on coverage and downstream
performance. Coverage is the fraction of training clips retained.
$\Delta$OD, $\Delta$MOT, and $\Delta$IS denote the change in
DETR/YOLOv8 AP, ByteTrack MOTA, and Mask2Former mIoU, respectively,
when replacing raw ADGV-Bench with the filtered subset.}
\label{tab:threshold_sweep}
\end{table}

The overall quality score $S_{\text{ADGVE}}$ is used to decide whether
a generated clip is retained for training. While the main paper adopts
a default threshold $\tau = 0.2$, here we explicitly study how the
choice of $\tau$ trades off coverage (i.e. how many frames survive) versus
downstream gains for detection, tracking, and instance segmentation.

We start from the ADGV-Bench training split and form three filtered
variants by thresholding the ADGVE score at
$\tau \in \{0.1, 0.2, 0.3\}$. For each threshold we measure
(i) coverage, defined as the fraction of clips whose scores exceed
$\tau$, and (ii) downstream performance when training the same
backbones as Sec.~4 on the corresponding filtered set and evaluating
on the ADGV-Bench test split. We reuse the detectors, trackers, and
segmenters from Tab.~5 (DETR and YOLOv8 for object detection, ByteTrack for multi object tracking, and Mask2Former for instance segmentation) and report the improvement over
training on raw ADGV-Bench.

Tab.~\ref{tab:threshold_sweep} summarizes the results. Lower thresholds
(e.g., $\tau = 0.1$) keep most clips but admit many low-quality videos,
leading to smaller gains over the raw baseline. Higher thresholds
(e.g., $\tau = 0.3$) produce very clean subsets but significantly
reduce coverage, which slightly hurts detectors and segmenters that
benefit from larger sample sizes. Consistent with Sec.~4.3, the
intermediate threshold $\tau = 0.2$ provides the best trade-off: it
retains roughly half of the generated clips while yielding the largest
average improvement across all four downstream models. This justifies
our choice of $\tau = 0.2$ as the default threshold throughout the
main paper.

\subsection{Human Annotation Confidence vs ADGVE Filtering}

\begin{table}[t]
  \centering
  \small
  \setlength{\tabcolsep}{4pt}
  \resizebox{\linewidth}{!}{
  \begin{tabular}{l|ccc|ccc}
    \toprule
    \multirow{2}{*}{Train Set} &
    \multicolumn{3}{c|}{DETR} &
    \multicolumn{3}{c}{YOLOv8} \\
    & AP & AP50 & AP75 & AP & AP50 & AP75 \\
    \midrule
    ADGV-bench-train          & 23.5 & 36.6 & 24.7 & 23.2 & 35.3 & 25.3 \\
    ADGV-bench-train-H        & 35.5 & 66.9 & 36.4 & 34.6 & 64.5 & 35.4 \\
    ADGV-bench-train-M        & 25.0 & 44.2 & 24.4 & 25.3 & 44.8 & 23.8 \\
    ADGV-bench-train-L        &  8.7 & 19.6 &  7.3 &  9.3 & 20.1 &  8.7 \\
    ADGV-bench-train-filtered & 33.2 & 53.6 & 33.5 & 30.2 & 52.1 & 31.8 \\
    \bottomrule
  \end{tabular}
  }
  \caption{Object detection results on ADGV-Bench. All models are
  evaluated on the ADGV-bench-test set using COCO-style metrics.
  Rows correspond to different training subsets based on human
  confidence labels (H/M/L) and the ADGVE-filtered clips. To isolate quality from data-volume effects, we always match the number of ADGV-Bench frames used in all train set.}
  \label{tab:od_ablation}
\end{table}


During ADGV-Bench object annotation, in addition to making basic bounding boxes with category, we also tag each bounding box with a discrete confidence label \{\texttt{high}, \texttt{medium}, \texttt{low}\} to reflect how clearly the object is rendered in the AI-generated video. 
These labels implicitly encode human judgements of geometric distortion, texture artifacts, and category ambiguity. 
We would like to understand how our automatic ADGVE filtering performance compares to this human, object-level notion of quality.

Starting from the raw ADGV-Bench training split, we construct four
variants:
\begin{itemize}[leftmargin=1.5em]
\item \textbf{ADGVE-bench-train}: original annotations with all
  boxes, regardless of confidence.
\item \textbf{ADGVE-bench-train-H/M/L}: subsets that retain only boxes
  whose confidence label is \texttt{high}, \texttt{medium}, or
  \texttt{low}, respectively (boxes with other labels are removed).
\item \textbf{ADGVE-bench-train-filtered}: the subset of clips whose
  ADGVE score exceeds the default threshold $\tau = 0.2$, with all
  boxes kept within those clips.
\end{itemize}
We then train DETR~\cite{DETR} and YOLOv8~\cite{YOLOv8} from scratch on each variant using the same hyperparameters as Sec.~4 and evaluate COCO-style AP, AP$_{50}$,
and AP$_{75}$ on the common ADGV-Bench test set. Results are reported in Tab.~\ref{tab:od_ablation}.

As it shown, training only on high-confidence boxes yields the best AP for both
DETR and YOLOv8. Using only medium-confidence boxes performs slightly better
than the full raw set, while the low-confidence subset leads to a
severe drop in performance, confirming that heavily distorted objects
are harmful for detector training.

Our ADGVE-filtered training sits between the “high-only’’ and “medium-only’’
settings: it clearly outperforms training on the raw or low-confidence
sets, and its AP is close to the high-confidence upper bound despite
operating purely at the clip level. 
This suggests that ADGVE’s automatically filtered subset captures a similar notion of useful, human-trustworthy training data as the annotators’ confidence labels.
In other words, ADGVE acts as a scalable, video-level proxy for human judgement of object quality, approaching the benefits of manually curated high-confidence annotations without requiring additional human effort at deployment time.

\end{document}